\documentclass[10pt,twocolumn,letterpaper]{article}

\usepackage{wacv}
\usepackage{times}
\usepackage{graphicx,import}
\usepackage{amsmath,amssymb} %
\usepackage{color}
\usepackage{microtype}
\usepackage{booktabs}
\usepackage{multirow}
\usepackage{graphbox}

\definecolor{rowblue}{RGB}{220,230,240}
\definecolor{myorchid}{RGB}{150,10,30}
\definecolor{myblue}{RGB}{10,30,250}
\definecolor{mygreen}{RGB}{10,190,10}

\newcommand{\myparagraph}[1]{\vspace{.2cm} \noindent \textbf{#1} \quad}

\wacvfinalcopy %

\ifwacvfinal
\usepackage[breaklinks=true,colorlinks,bookmarks=false]{hyperref}
\else
\usepackage[pagebackref=true,breaklinks=true,colorlinks,bookmarks=false]{hyperref}
\fi

\begin{document}

\title{Improved Techniques for Training Single-Image GANs}

\author{Tobias Hinz$^1$, Matthew Fisher$^2$, Oliver Wang$^2$, and Stefan Wermter$^1$\\
$^1$Knowledge Technology, University of Hamburg, Germany\\
$^2$Adobe Research
}

\maketitle
\thispagestyle{empty}

\def\reduceheight{-0.25em}

\begin{abstract}
Recently there has been an interest in the potential of learning generative models from a \emph{single} image, as opposed to from a large dataset. 
This task is of significance, as it means that generative models can be used in domains where collecting a large dataset is not feasible. 
However, training a model capable of generating realistic images from only a single sample is a difficult problem. 
In this work, we conduct a number of experiments to understand the challenges of training these methods and propose some best practices that we found allowed us to generate improved results over previous work.
One key piece is that, unlike prior single image generation methods, we \emph{concurrently} train several stages in a sequential multi-stage manner, allowing us to learn models with fewer stages of increasing image resolution.
Compared to a recent state of the art baseline, our model is up to six times faster to train, has fewer parameters, and can better capture the global structure of images. 
\end{abstract}

\section{Introduction}
\vspace{\reduceheight}
Generative Adversarial Networks (GANs) \cite{goodfellow2014generative} are capable of generating realistic images \cite{brock2018large} that are often indistinguishable from real ones \cite{karras2019analyzing}.
The resulting models can be used for different tasks, such as unconditional and conditional image synthesis \cite{karras2019style,hinz2019generating}, image inpainting \cite{demir2018patch}, and image-to-image translation \cite{isola2017image,zhu2017unpaired}.
However, most GANs are trained on large datasets, typically consisting of tens of thousands of images which can be time-consuming and expensive.
In some cases, it might be preferable to train a generative model on a small number of images or, in the limit, on a single image.
This is useful if we want to obtain variations of a given image, work with a very specific image or style, or only have access to little training data.
The recently proposed SinGAN~\cite{shaham2019singan} introduces a GAN that is trained on a single image for tasks such as unconditional image generation and harmonization.

SinGAN is trained in a multi-stage and multi-resolution approach, where the training starts at a very low resolution (e.g.\ $25 \times 25$ pixels) at the first stage.
The training progresses through several ``stages'', at each of which more layers are added to the generator and the image resolution is increased.
At each stage all previously trained stages (i.e.\ the generator's lower layers) are frozen and only the newly added layers are trained.
We find that exactly how multi-stage and multi-resolution training is handled is critical. 
In particular, training only one stage at a given time limits interactions between different stages, and propagating images instead of feature maps from one generator stage to the next negatively affects the learning process.
Conversely, training all stages end-to-end causes overfitting in the single image scenario, where the network collapses to generating only the input image.
We experiment with this balance, and find a promising compromise, training multiple stages in parallel with decreased learning rates, and find that this improves the learning process, leading to more realistic images with less training time.
Furthermore, we show how it is possible to directly trade-off image quality for image variance, where training more stages in parallel means a higher global image consistency at the price of less variation.

We also conduct experiments over the choice of rescaling parameters, i.e.\ how we decide at which image resolution to train at each stage.
We observe that the quality of the generated images, especially the overall image layout, quickly degrades when there are not enough training stages with small resolution.
Our experiments show that lower stages with smaller resolutions are important for the overall image layout, while higher stages with larger resolution are important for the final image texture and color.
We find that we only need relatively few training stages with high-resolution images in order to still generate images with the correct texture.
As a consequence, we put a higher weight on smaller resolution images during training while using fewer of the stages to train on high-resolution images.

\begin{figure*}[t]
    \centering
    \includegraphics[width=.9\textwidth]{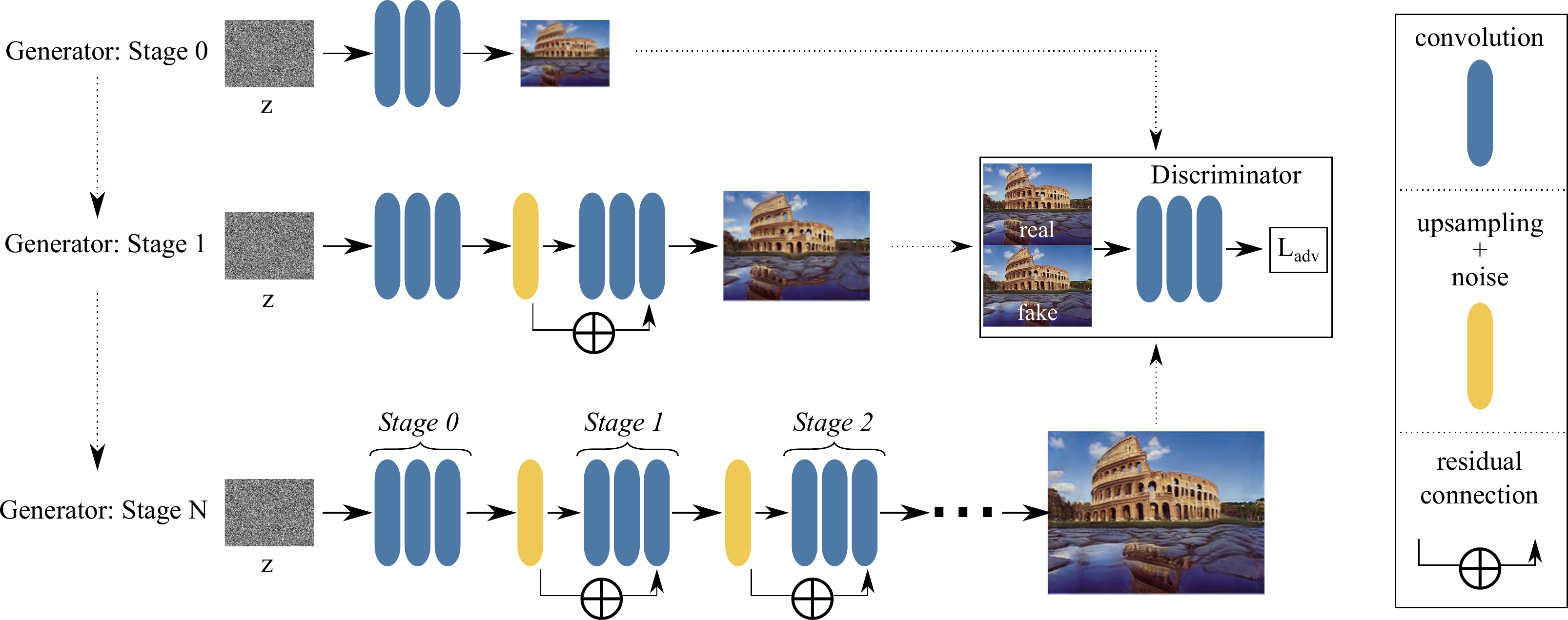}
    \caption{Overview of our model (ConSinGAN). We start training at `Stage 0' with a small generator and small image resolution. With increasing number of stages both the generator capacity and image resolution increase.}
    \label{fig:model}
\end{figure*}

Finally, since our model trains several stages in parallel, we can introduce a \textit{task-specific fine-tuning stage} which can be performed on any trained model.
For several tasks we show how to fine-tune the trained model on a given specific image to further improve results.
This shows benefits with as few as 500 additional training iterations and is, therefore, very fast (less than two minutes on our hardware).

Combining these proposed architecture and training modifications enables us to generate realistic images with fewer stages and significantly reduced overall training time (20-25 minutes versus 120-150 minutes in the original SinGAN work). 
To summarize, our main contributions are:
\begin{enumerate}
    \item We train several stages in parallel with different learning rates and can trade-off the variance in generated images vs. their conformity to the original training image.
    \item We do not generate images at intermediate stages but propagate features directly from one stage to the next.
    \item We improve the rescaling approach for multi-stage training, which enables us to train on fewer stages.
    \item We introduce a fine-tuning phase which can be used on pre-trained models to obtain optimal results for specific images and tasks.
\end{enumerate}

\begin{figure*}[h]
    \centering
    \renewcommand{\arraystretch}{1}
    \def\qualheight{7.5cm}
    \def\fs{\footnotesize}
    \begin{tabular}{*{1}{c@{\hspace{2.5px}}}p{3px}p{0px}*{1}{c@{\hspace{0.75px}}}}
    
    \fs Input &&& \fs Random Samples \\

    \includegraphics[align=c,height=\qualheight]{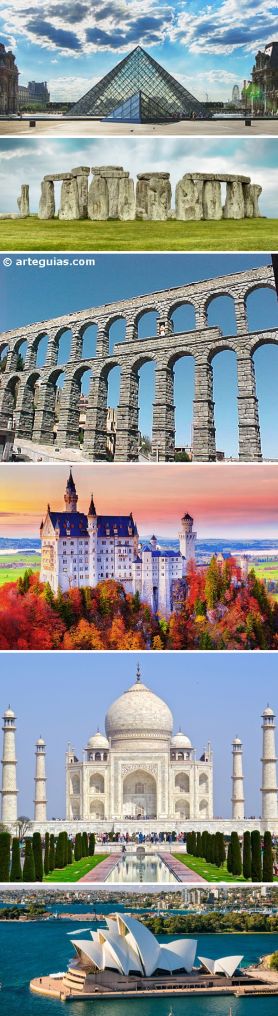}&\multicolumn{1}{l|}{}&&
    \includegraphics[align=c,height=\qualheight]{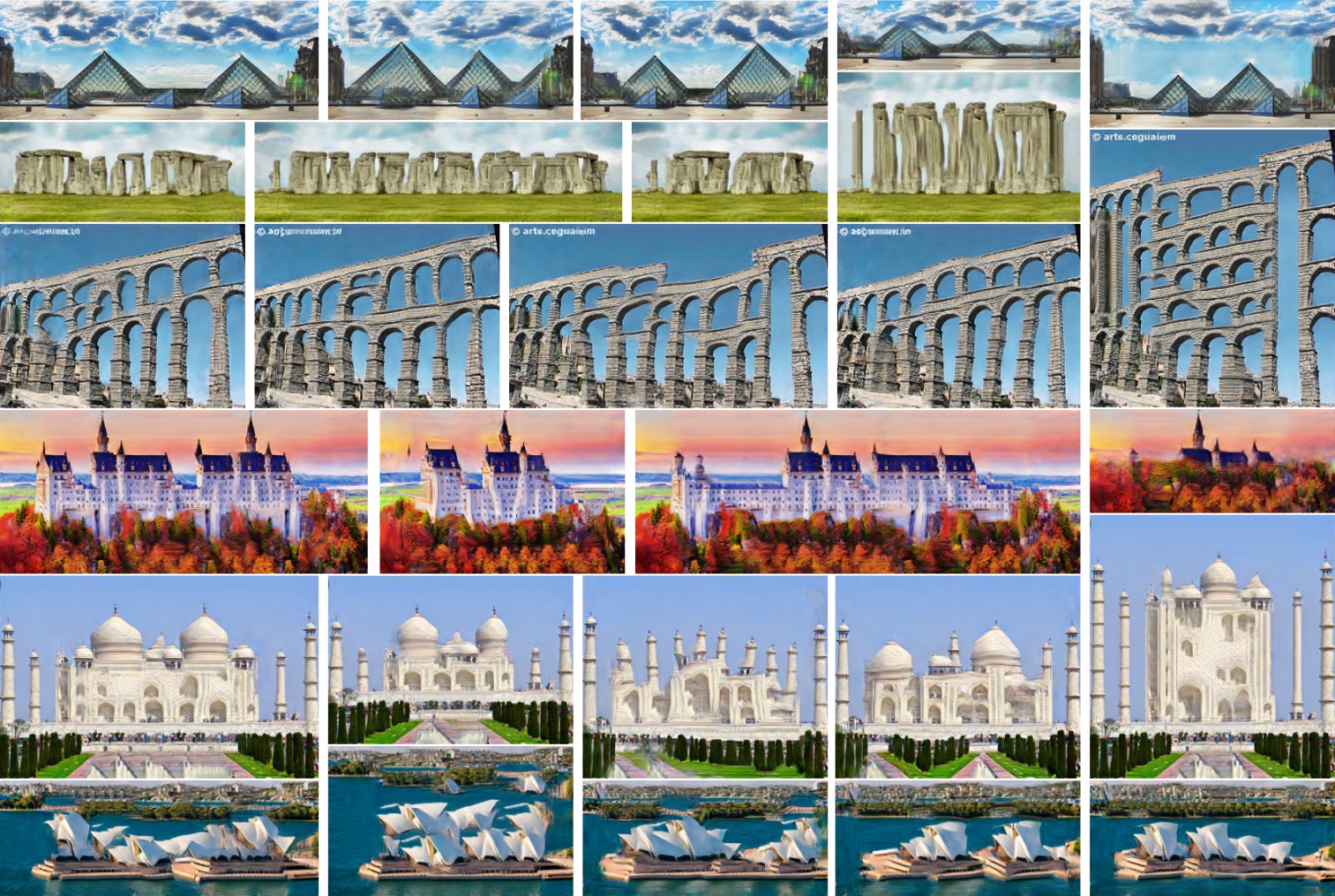} \\
    
    \end{tabular}
    \vspace{0.5em}
    \caption{Example of unconditionally generated images showing complex global structure generated by ConSinGAN.}
    \label{fig:unconditional:consingan}
\end{figure*}

\section{Related Work}
\vspace{\reduceheight}
Learning the statistics and distribution of patches of a single image has been known to provide a powerful prior since the empirical entropy of patches inside a single image is smaller than the empirical entropy of patches inside a distribution of images \cite{zontak2011internal}.
By using this prior, many tasks such as inpainting \cite{ulyanov2018deep,zhang2019internal}, denoising \cite{zontak2013separating}, deblurring \cite{michaeli2014blind}, retargeting \cite{mastan2019multi,mastan2020dcil}, and segmentation \cite{gandelsman2019double} can be solved with only a single image.
In particular, image super-resolution \cite{yang2019deep,huang2015single,glasner2009super,shocher2018zero,bell2019blind} and editing \cite{cho2008patch,dekel2015revealing,he2012statistics,mechrez2019saliency,tlusty2018modifying,mao2019program} from a single image have been shown to be successful and a large body of work focuses specifically on this task.
Recent work also shows that training a model on a single image with self-supervision and data augmentation can be enough to learn powerful feature extraction layers \cite{asano2020critical}.

Approaches that train GAN models on single images are still relatively rare and are usually based on a bidirectional similarity measure for image summarization~\cite{simakov2008summarizing}.
Some approaches do not use natural images, but instead train only on texture images \cite{jetchev2016texture,zhou2018non,bergmann2017learning,li2016precomputed}.
At this time, only few models are capable of being trained on a single `natural' image \cite{shaham2019singan, Shocher_2019_ICCV, vinker2020deep}.
Other novel approaches target applications such as image-to-image translation with only two images as training data \cite{lin2020tuigan, benaim2020structural}.

The work most relevant to our approach is SinGAN \cite{shaham2019singan} which is the only model that can perform unconditional image generation after being trained on a single natural image.
SinGAN trains both the generator and the discriminator over multiple stages of different image resolutions as it is useful to learn statistics of image patches across different image scales \cite{bagon2008good}.
The output at each stage is an image which is used as input to the next stage and each stage is trained individually while the previous stages are kept frozen.

\begin{figure*}[t]
    \centering
    \renewcommand{\arraystretch}{1.0}
    \def\qualheight{1.0cm}
    \def\fsv{\tiny}
    \def\fsh{\footnotesize}
    
    \begin{tabular}{*{6}{c@{\hspace{1px}}}p{1px}*{6}{c@{\hspace{1px}}}}
    \multicolumn{6}{c}{\fsh Number of Concurrently Trained Stages} && \multicolumn{6}{c}{\fsh Number of Concurrently Trained Stages} \\
    \fsh 1 & \fsh 2 & \fsh 3 & \fsh 4 & \fsh 5 & \fsh 6 && \fsh 1 & \fsh 2 & \fsh 3 & \fsh 4 & \fsh 5 & \fsh 6 \\
    
    \includegraphics[align=c,height=\qualheight]{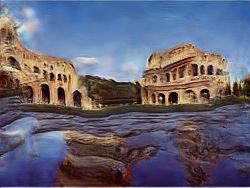}&
    \includegraphics[align=c,height=\qualheight]{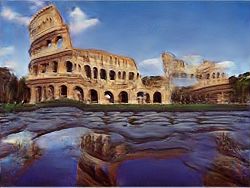}&
    \includegraphics[align=c,height=\qualheight]{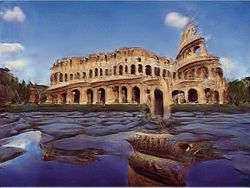}&
    \includegraphics[align=c,height=\qualheight]{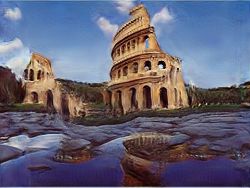}&
    \includegraphics[align=c,height=\qualheight]{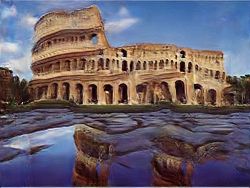}&
    \includegraphics[align=c,height=\qualheight]{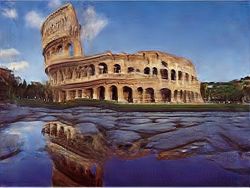}&&
    \includegraphics[align=c,height=\qualheight]{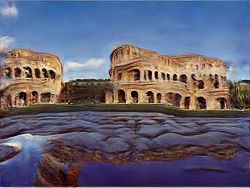}&
    \includegraphics[align=c,height=\qualheight]{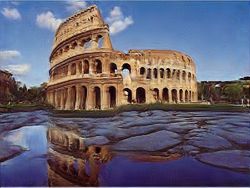}&
    \includegraphics[align=c,height=\qualheight]{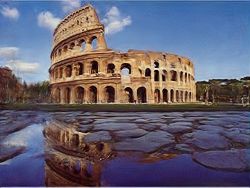}&
    \includegraphics[align=c,height=\qualheight]{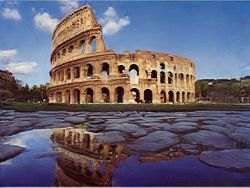}&
    \includegraphics[align=c,height=\qualheight]{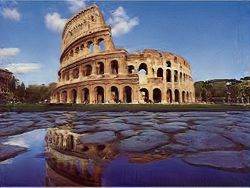}&
    \includegraphics[align=c,height=\qualheight]{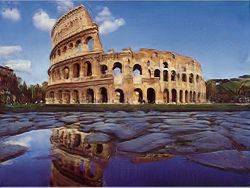}\\
    
    \includegraphics[align=c,height=\qualheight]{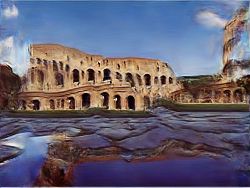}&
    \includegraphics[align=c,height=\qualheight]{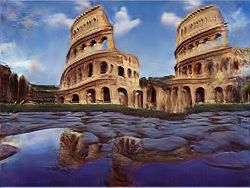}&
    \includegraphics[align=c,height=\qualheight]{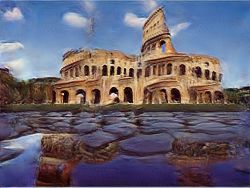}&
    \includegraphics[align=c,height=\qualheight]{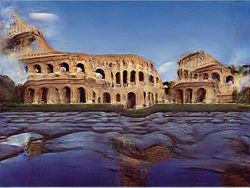}&
    \includegraphics[align=c,height=\qualheight]{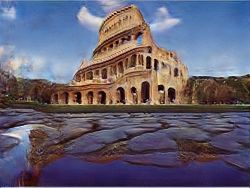}&
    \includegraphics[align=c,height=\qualheight]{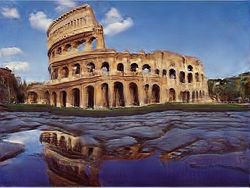}&&
    \includegraphics[align=c,height=\qualheight]{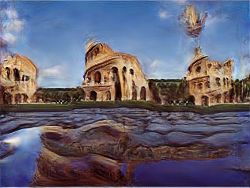}&
    \includegraphics[align=c,height=\qualheight]{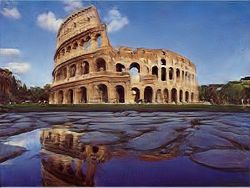}&
    \includegraphics[align=c,height=\qualheight]{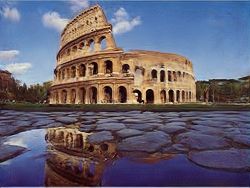}&
    \includegraphics[align=c,height=\qualheight]{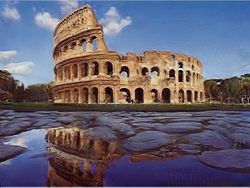}&
    \includegraphics[align=c,height=\qualheight]{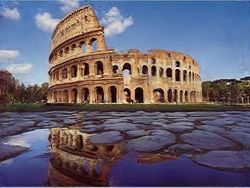}&
    \includegraphics[align=c,height=\qualheight]{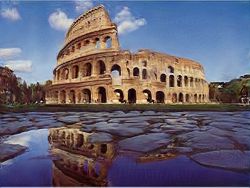}\\
    
    \includegraphics[align=c,height=\qualheight]{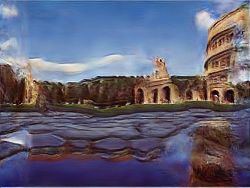}&
    \includegraphics[align=c,height=\qualheight]{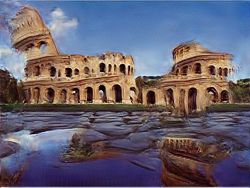}&
    \includegraphics[align=c,height=\qualheight]{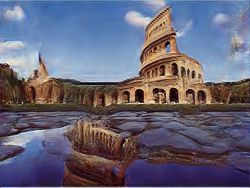}&
    \includegraphics[align=c,height=\qualheight]{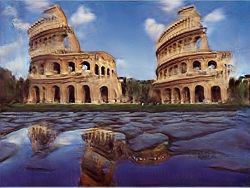}&
    \includegraphics[align=c,height=\qualheight]{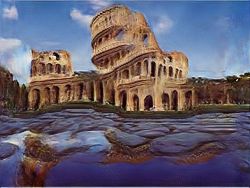}&
    \includegraphics[align=c,height=\qualheight]{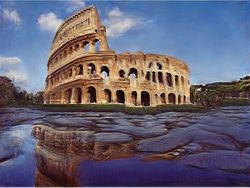}&&
    \includegraphics[align=c,height=\qualheight]{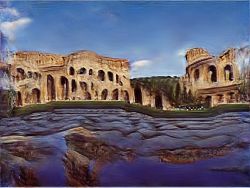}&
    \includegraphics[align=c,height=\qualheight]{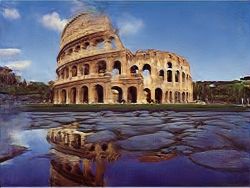}&
    \includegraphics[align=c,height=\qualheight]{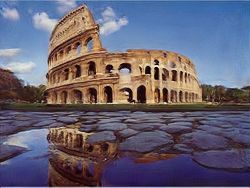}&
    \includegraphics[align=c,height=\qualheight]{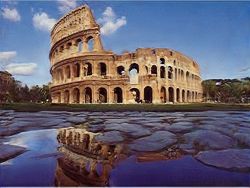}&
    \includegraphics[align=c,height=\qualheight]{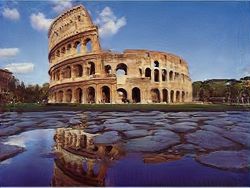}&
    \includegraphics[align=c,height=\qualheight]{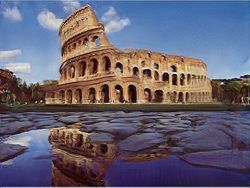}\\

    \multicolumn{6}{c}{\fsh$\delta = 0.1$} && \multicolumn{6}{c}{\fsh $\delta = 0.5$} \\
    \end{tabular}

    \caption{Effect of learning rate scale $\delta$ and concurrently trained stages for a model with six stages. Images are randomly selected.}
    \label{fig:overfitting}
\end{figure*}
\begin{figure*}
    \centering
    \renewcommand{\arraystretch}{1.0}
    \def\qualheight{0.95cm}
    \def\fsv{\tiny}
    \def\fsh{\footnotesize}
    \def\sp{\hspace{0.0cm}}
    \hspace{0.0cm}
    \parbox{.46\linewidth}{\begin{tabular}{cc*{3}{c@{\hspace{14px}}}}
    & \fsh Input & \multicolumn{3}{c}{\fsh Generated Images}  \\
    \rotatebox[origin=c]{90}{\fsv{$\delta=0.5$}} & 
    \multicolumn{1}{l|}{\includegraphics[align=c,height=\qualheight]{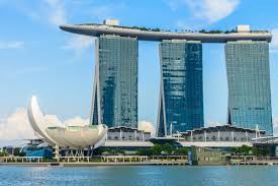}}&
    \includegraphics[align=c,height=\qualheight]{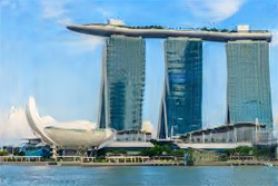} &
    \includegraphics[align=c,height=\qualheight]{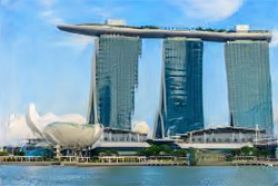} &
    \includegraphics[align=c,height=\qualheight]{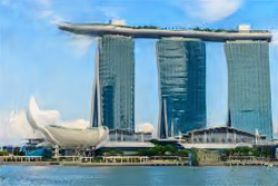}  \\
    \rotatebox[origin=c]{90}{\fsv{$\delta=0.1$}} &
    \multicolumn{1}{l|}{\includegraphics[align=c,height=\qualheight]{figures/unconditional-lr-effect/single-images/0_training.jpg}}&
    \includegraphics[align=c,height=\qualheight]{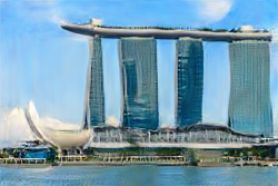}&
    \includegraphics[align=c,height=\qualheight]{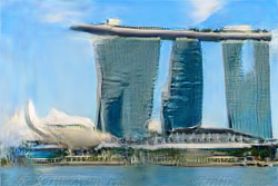}&
    \includegraphics[align=c,height=\qualheight]{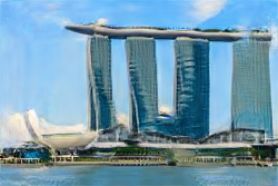}
    \end{tabular}}
    \hspace{0.5cm}
    \parbox{.45\linewidth}{\begin{tabular}{c*{3}{c@{\hspace{2px}}}}
    \fsh Input & \multicolumn{3}{c}{\fsh Generated Images}  \\
    \multicolumn{1}{l|}{\includegraphics[align=c,height=\qualheight]{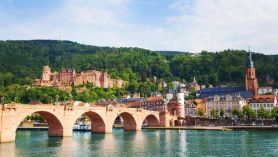}} &
    \includegraphics[align=c,height=\qualheight]{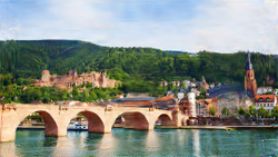}&
    \includegraphics[align=c,height=\qualheight]{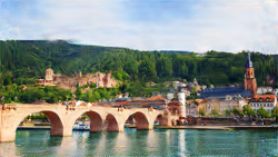}&
    \includegraphics[align=c,height=\qualheight]{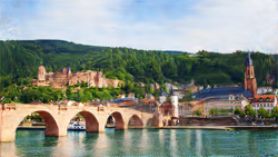}\\
    \multicolumn{1}{l|}{\includegraphics[align=c,height=\qualheight]{figures/unconditional-lr-effect/single-images/2_training.jpg}} &
    \includegraphics[align=c,height=\qualheight]{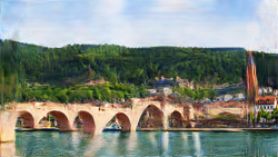}&
    \includegraphics[align=c,height=\qualheight]{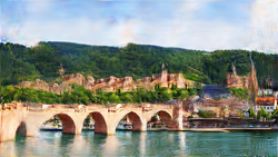}&
    \includegraphics[align=c,height=\qualheight]{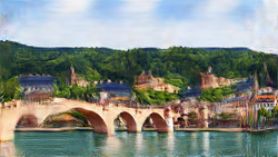}\\
    \end{tabular}}
    
    \def\qualheight{1.55cm}
    \hspace{0.0cm}
    \parbox{.46\linewidth}{\begin{tabular}{cp{0.1mm}c*{5}{c@{\hspace{2px}}}}
    \rotatebox[origin=c]{90}{\fsv{$\delta=0.5$}} & &
    \multicolumn{1}{l|}{\includegraphics[align=c,height=\qualheight]{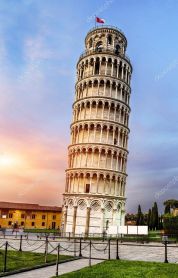}}&
    \includegraphics[align=c,height=\qualheight]{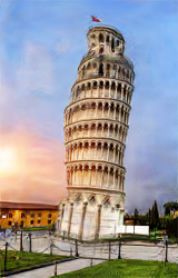}&
    \includegraphics[align=c,height=\qualheight]{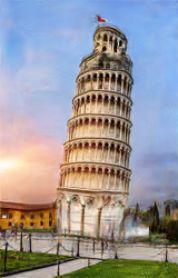}&
    \includegraphics[align=c,height=\qualheight]{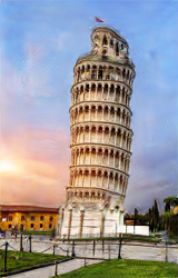}&
    \includegraphics[align=c,height=\qualheight]{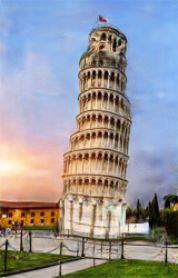}&
    \includegraphics[align=c,height=\qualheight]{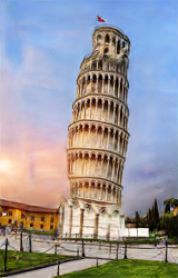}\\
    \rotatebox[origin=c]{90}{\fsv{$\delta=0.1$}} & & \multicolumn{1}{l|}{\includegraphics[align=c,height=\qualheight]{figures/unconditional-lr-effect/single-images/1_training.jpg}}&
    \includegraphics[align=c,height=\qualheight]{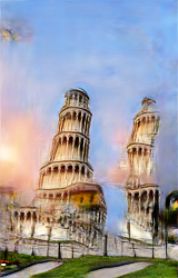}&
    \includegraphics[align=c,height=\qualheight]{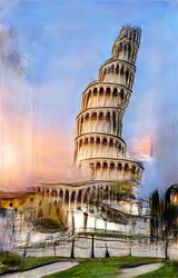}&
    \includegraphics[align=c,height=\qualheight]{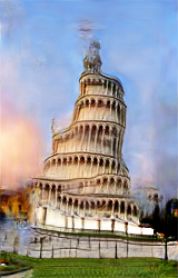}&
    \includegraphics[align=c,height=\qualheight]{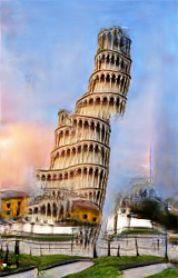}&
    \includegraphics[align=c,height=\qualheight]{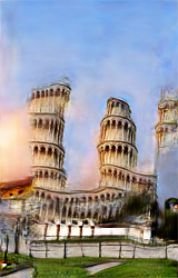}\\
    \end{tabular}}
    \hspace{0.5cm}
    \parbox{.45\linewidth}{\begin{tabular}{p{1mm}c*{4}{c@{\hspace{5px}}}}
    & \multicolumn{1}{l|}{\includegraphics[align=c,height=\qualheight]{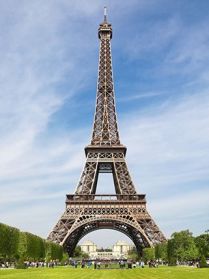}} &
    \includegraphics[align=c,height=\qualheight]{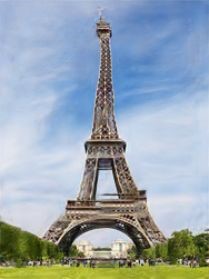}&
    \includegraphics[align=c,height=\qualheight]{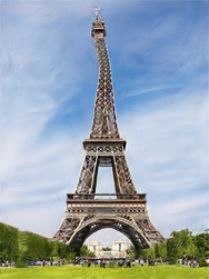}&
    \includegraphics[align=c,height=\qualheight]{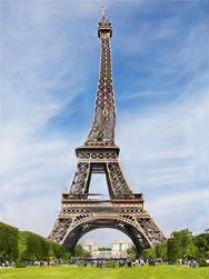}&
    \includegraphics[align=c,height=\qualheight]{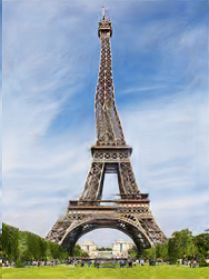}\\
    
    & \multicolumn{1}{l|}{\includegraphics[align=c,height=\qualheight]{figures/unconditional-lr-effect/single-images/3_training.jpg}}&
    \includegraphics[align=c,height=\qualheight]{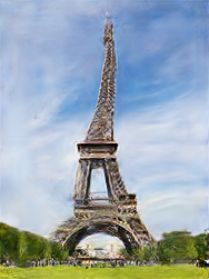}&
    \includegraphics[align=c,height=\qualheight]{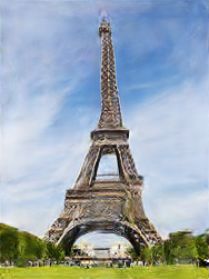}&
    \includegraphics[align=c,height=\qualheight]{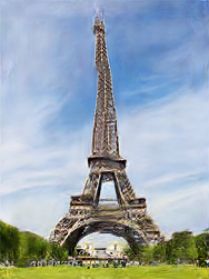}&
    \includegraphics[align=c,height=\qualheight]{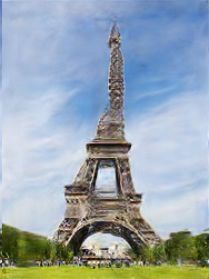}\\
    \end{tabular}}
    
    \vspace{0.5em}
    \caption{Effect of the learning rate scale $\delta$ during training of ConSinGAN.}
    \label{fig:unconditional:delta}
\end{figure*}

\section{Methodology}
\vspace{\reduceheight}
\label{sec:methodology}

We now describe our findings in more detail, starting with the training of a multi-stage architecture, followed by best practices we found for scaling learning rate and image resolutions at different stages during training. 

\myparagraph{Multi-stage Training}
Multi-scale image generation is of critical importance \cite{shaham2019singan}, however, there are many ways in which this can be realized.  
SinGAN only trains the current (highest) stage of its generator and freezes the parameters of all previous stages.
ProGAN \cite{karras2017progressive} presents a progressive growing scheme that adds levels with all weights unfrozen, and more recently~\cite{Karnewar_2019_CVPR,karras2019style} train the entire pyramid jointly. 

In this work, we investigate whether the model can be trained end-to-end, rather than with training being fixed at intermediate stages, even in the single image task.
However, we find that training all stages leads to \emph{overfitting} (see \autoref{fig:overfitting}), i.e.\ the generator only generates the original training image without any variation.
We develop a novel progressive growing technique that trains multiple, \emph{but not all}, stages concurrently while simultaneously using progressively smaller learning rates at lower stages.
Since we train several stages of our model concurrently for a single image we refer to our model as `Concurrent-Single-Image-GAN' (ConSinGAN).

Training ConSinGAN starts on a coarse resolution for a number of iterations, learning a mapping from a random noise vector $z$ to a low-resolution image (see ``Generator: Stage 0'' in \autoref{fig:model}).
Once training of stage $n$ has converged, we increase the size of our generator by adding three additional convolutional layers.
In contrast to SinGAN, each stage gets the raw features from the previous stage as input, and previous layers are not fixed.
We add a residual connection  \cite{he2016deep} from the original features to the output of the newly added convolutional layers (see ``Generator: Stage 1'' in \autoref{fig:model}).
We repeat this process $N$ times until we reach our desired output resolution.
We add additional noise to the features at each stage \cite{isola2017image,zhu2017toward} to improve diversity.
In our default setting, we jointly train the last three stages of a generator (see ``Generator: Stage N'' in \autoref{fig:model}).
While it is possible to train more than three stages concurrently, we observed that this rapidly leads to severe overfitting (\autoref{fig:overfitting}).

We use the same patch discriminator \cite{isola2017image} architecture and loss function as the original SinGAN.
This means that the receptive field in relation to the size of the generated image gets smaller as the number of stages increases, meaning that the discriminator focuses more on global layout at lower resolutions and more on texture at higher resolutions.
In contrast to SinGAN we do not increase the capacity of the discriminator at higher stages, but use the same number of parameters at every stage.
We initialize the discriminator for a given stage $n$ with the weights of the discriminator of the previous stage $n-1$ at all stages.
At a given stage $n$, we optimize the sum of an adversarial and a reconstruction loss:
\begin{equation}
    \underset{G_n}{min}\ \underset{D_n}{max}\ \mathcal{L}_{\textit{adv}}(G_n, D_n) + \alpha \mathcal{L}_{\textit{rec}}(G_n).
\end{equation}
$\mathcal{L}_{\textit{adv}}(G_n, D_n)$ is the WGAN-GP adversarial loss~\cite{gulrajani2017improved}, while the reconstruction loss is used to improve training stability ($\alpha=10$ for all our experiments).
For the reconstruction loss the generator $G_n$ gets as input a downsampled version ($x_0$) of the original image ($x_N$) and is trained to reconstruct the image at the given resolution of stage $n$:
\begin{equation}
    \mathcal{L}_{\textit{rec}}(G_n) = \vert\vert G_n(x_0) - x_n \vert\vert_2^2.
\end{equation}
The discriminator is always trained in the same way, i.e.\ it gets as input either a generated or a real image and is trained to maximise $\mathcal{L}_{\textit{adv}}$.
Our generator, however, is trained slightly differently depending on the final task.

\myparagraph{Task Specific Generator Training}
For each task we use the original image $x_n$ for the reconstruction loss $\mathcal{L}_{\textit{rec}}$.
The input for the adversarial loss $\mathcal{L}_{\textit{adv}}$, however, depends on the task.
For unconditional image generation the input to the generator is simply a randomly sampled noise vector for $\mathcal{L}_{\textit{adv}}$.
However, we found that if the desired task is known beforehand, better results can be achieved by training with a different input format. 
For example, for image harmonization, we can instead train using the original image with augmentation transformations applied as input.
The intuition for this is that a model that is used for image harmonization does not need to learn how to generate realistic images from random noise, but rather should learn how to harmonize different objects and color distributions.
To simulate this task, we apply random combinations of augmentation techniques such as additive noise and color transforms to the original image $x_N$ at each iteration.
The generator gets the augmented image as input and needs to transform it back to an image that should resemble the original distribution.

\myparagraph{Learning Rate Scaling} 
The space of all learning rates for each stage is large and has a big impact on the final image quality. 
At any given stage $n$, we found that instead of training all stages ($n$, $n-1$, $n-2$, $...$) with the same learning rate, using a lower learning rate on earlier stages ($n-1$, $n-2$, $...$) helps reduce overfitting.
If the learning rate at lower stages is too large (or too many stages are trained concurrently), the model generator quickly collapses and only generates the training image (\autoref{fig:overfitting}).
Therefore, we propose to scale the learning rate $\eta$ with a factor $\delta$.
This means that for generator $G_n$ stage $n$ is trained with learning rate $\delta^0\eta$, stage $n-1$ is trained with a learning rate $\delta^1\eta$, stage $n-2$ with $\delta^2\eta$, etc.
In our experiments, we found that setting $\delta=0.1$ gives a good trade-off between image fidelity and diversity (see \autoref{fig:overfitting} and \autoref{fig:unconditional:delta}).

\myparagraph{Improved Image Rescaling}
Another critical design choice is around what kind of multiscale pyramid to use. 
SinGAN originally proposes to downsample the image $x$ by a factor of $r^{N-n}$ for each stage $n$ where $r$ is a scalar with default value $0.75$.
As a result, SinGAN is usually trained on eight to ten stages for a resolution of $250$ width or height.
When the images are downsampled more aggressively (e.g. $r=0.5$) fewer stages are needed, but the generated images lose much of their global coherence.

We observe that this is the case when there are not enough stages at low resolution (roughly fewer than $60$ pixels at the longer side).
When training on images with a high resolution, the global layout is already ``decided'' and only texture information is important since the discriminator's receptive field is always $11\times 11$.
To achieve a certain global image layout we need a certain number of stages (usually at least three) at low resolution, but we do not need many stages a high resolution.
We adapt the rescaling to not be strictly geometric (i.e.\ $x_n = x_0\times r^{N-n}$), but instead to keep the density of low-resolution stages higher than the density of high-resolution stages:
\begin{equation}
    x_n = x_N\times r^{((N-1)/\textit{log}(N))*\textit{log}(N-n)+1} \text{\ for\ } n=0, ..., N-1
    \label{formula:rescaling}
\end{equation}
For example, with a rescaling scalar $r=0.55$ we get six stages with the following resolutions and we observe that our new rescaling approach (second line) has more stages with smaller resolutions compared to the original rescaling approach (first line):
\begin{align*}
    25\!&\times\!34, 38\!\times\!50, 57\!\times\!75, 84\!\times\!112, 126\!\times\!167, 188\!\times\!250,\\
    25\!&\times\!34, 32\!\times\!42, 42\!\times\!56, 63\!\times\!84, \hspace{0.5em}126\!\times\!167, 188\!\times\!250.
\end{align*}
To summarize our main findings, we produce feature maps rather than images at each stage, we train multiple stages concurrently, we propose a modified rescaling pyramid, and we present a task-specific training variation. 

\begin{figure}
    \centering
    \renewcommand{\arraystretch}{1.0}
    \def\fsv{\tiny}
    \def\fsh{\footnotesize}
    \def\qualheight{0.9cm}
    \begin{tabular}{ccc*{2}{c@{\hspace{2px}}}}
    && \fsh{Input} & \multicolumn{2}{c}{\fsh{Generated Images}} \\
    
    \rotatebox[origin=c]{90}{\fsv{SinGAN}} & 
    \rotatebox[origin=c]{90}{\fsv{8 Stages}} &
    \multicolumn{1}{l|}{\includegraphics[align=c,height=\qualheight]{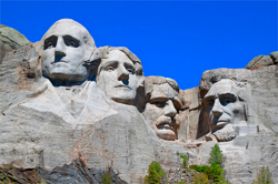}}&
    \includegraphics[align=c,height=\qualheight]{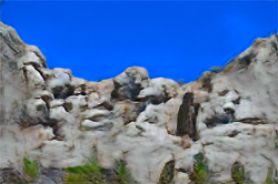}&
    \includegraphics[align=c,height=\qualheight]{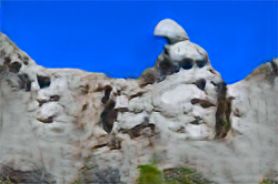}\\
    \rotatebox[origin=c]{90}{\fsv{SinGAN}} & 
    \rotatebox[origin=c]{90}{\fsv{6 Stages}} &
    \multicolumn{1}{l|}{\includegraphics[align=c,height=\qualheight]{figures/unconditional-comparison-singan/single-images/0_training.jpg}}&
    \includegraphics[align=c,height=\qualheight]{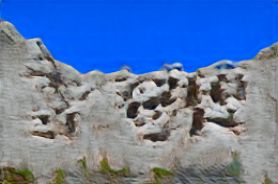}&
    \includegraphics[align=c,height=\qualheight]{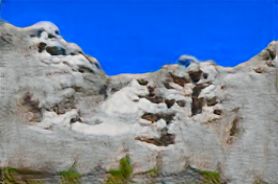}\\
    \rotatebox[origin=c]{90}{\fsv{ConSinGAN}} & 
    \rotatebox[origin=c]{90}{\fsv{6 Stages}} &
    \multicolumn{1}{l|}{\includegraphics[align=c,height=\qualheight]{figures/unconditional-comparison-singan/single-images/0_training.jpg}}&
    \includegraphics[align=c,height=\qualheight]{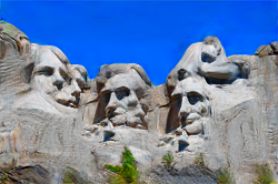}&
    \includegraphics[align=c,height=\qualheight]{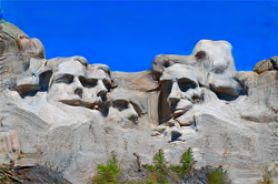}\\
    \end{tabular}
    
    \def\qualheight{1.3549cm}
    \begin{tabular}{ccc*{3}{c@{\hspace{2px}}}}
    \rotatebox[origin=c]{90}{\fsv{SinGAN}} & 
    \rotatebox[origin=c]{90}{\fsv{10 Stages}} &
    \multicolumn{1}{l|}{\includegraphics[align=c,height=\qualheight]{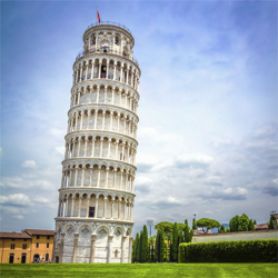}}&
    \includegraphics[align=c,height=\qualheight]{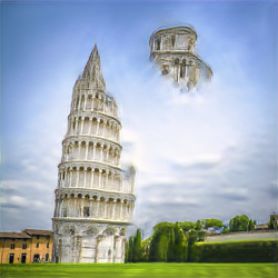}&
    \includegraphics[align=c,height=\qualheight]{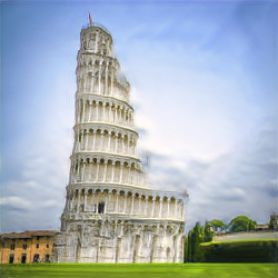}\\
    \rotatebox[origin=c]{90}{\fsv{SinGAN}} & 
    \rotatebox[origin=c]{90}{\fsv{5 Stages}} &
    \multicolumn{1}{l|}{\includegraphics[align=c,height=\qualheight]{figures/unconditional-comparison-singan/single-images/1_training.jpg}}&
    \includegraphics[align=c,height=\qualheight]{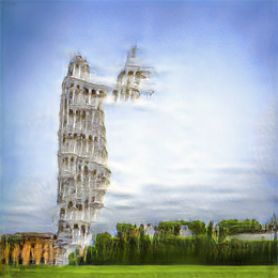}&
    \includegraphics[align=c,height=\qualheight]{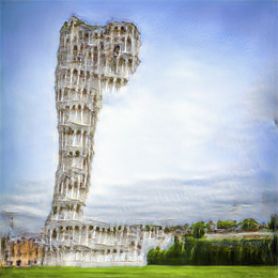}\\
    \rotatebox[origin=c]{90}{\fsv{ConSinGAN}} & 
    \rotatebox[origin=c]{90}{\fsv{5 Stages}} &
    \multicolumn{1}{l|}{\includegraphics[align=c,height=\qualheight]{figures/unconditional-comparison-singan/single-images/1_training.jpg}}&
    \includegraphics[align=c,height=\qualheight]{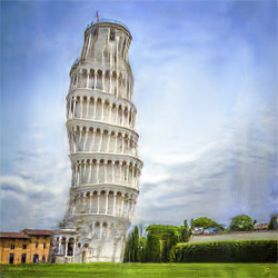}&
    \includegraphics[align=c,height=\qualheight]{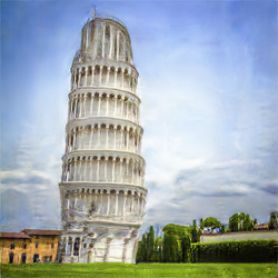}\\
    \end{tabular}
    
    \vspace{0.5em}
    \caption{Comparison of SinGAN and ConSinGAN.}
    \label{fig:unconditional:consingan:singan}
\end{figure}

\section{Results}
\vspace{\reduceheight}
We evaluate ConSinGAN on unconditional image generation and image harmonization in detail.\footnote{Code: \url{https://github.com/tohinz/ConSinGAN}}
For space reasons we focus on these two applications but note that other applications are also possible with ConSinGAN.
We show examples of other tasks such as image retargeting, editing, and animation in the supplementary material.

\begin{figure*}
    \centering
    \renewcommand{\arraystretch}{1}
    \def\qualheight{1.0cm}
    \def\fsh{\footnotesize}
    \def\fsv{\tiny}
    \begin{tabular}{c*{5}{c@{\hspace{1px}}}p{2px}*{5}{c@{\hspace{1px}}}}
    & \multicolumn{5}{c}{\fsh ConSinGAN\hspace{0.1em}-\hspace{0.1em}Number of Trained Stages} && \multicolumn{5}{c}{\fsh SinGAN \cite{shaham2019singan}\hspace{0.1em}-\hspace{0.1em}Number of Trained Stages}\\
    & \fsh 4 & \fsh 5 & \fsh 6 & \fsh 7 & \fsh 8 && \fsh 4 & \fsh 5 & \fsh 6 & \fsh 7 & \fsh 8 \\
    \multirow{3}[10]{*}{\rotatebox[origin=c]{90}{\fsv New Rescaling}} &
    \includegraphics[align=c,height=\qualheight]{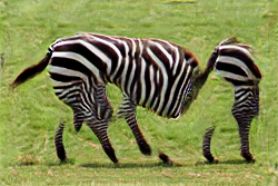}&
    \includegraphics[align=c,height=\qualheight]{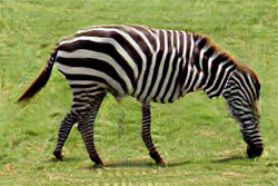}&
    \includegraphics[align=c,height=\qualheight]{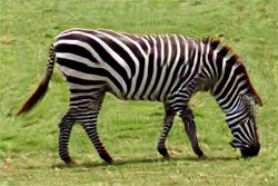}&
    \includegraphics[align=c,height=\qualheight]{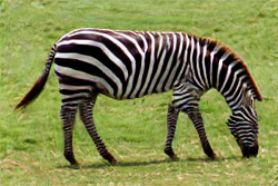}&
    \includegraphics[align=c,height=\qualheight]{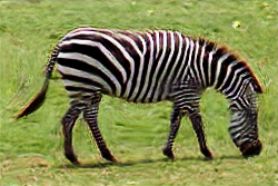}&&
    
    \includegraphics[align=c,height=\qualheight]{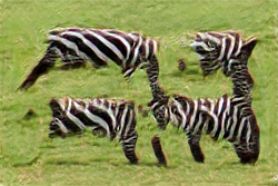}&
    \includegraphics[align=c,height=\qualheight]{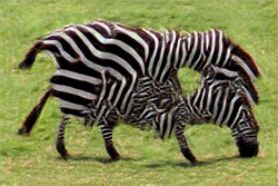}&
    \includegraphics[align=c,height=\qualheight]{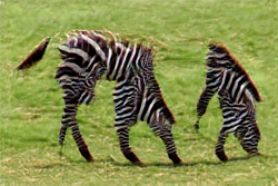}&
    \includegraphics[align=c,height=\qualheight]{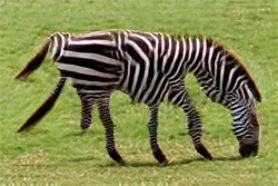}&
    \includegraphics[align=c,height=\qualheight]{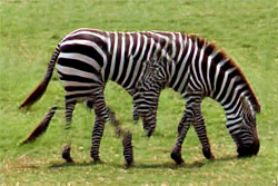}\\
    
    & \includegraphics[align=c,height=\qualheight]{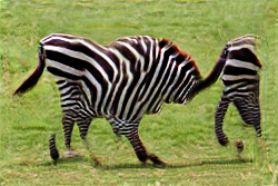}&
    \includegraphics[align=c,height=\qualheight]{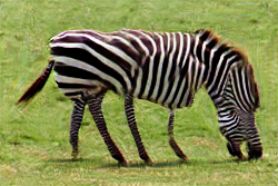}&
    \includegraphics[align=c,height=\qualheight]{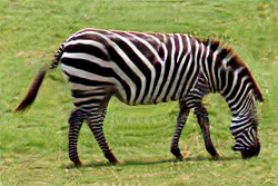}&
    \includegraphics[align=c,height=\qualheight]{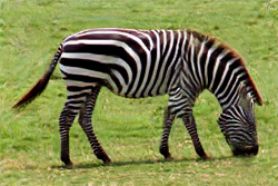}&
    \includegraphics[align=c,height=\qualheight]{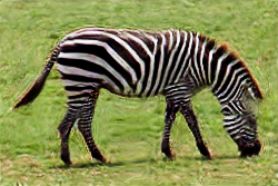}&&
    
    \includegraphics[align=c,height=\qualheight]{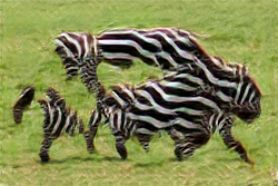}&
    \includegraphics[align=c,height=\qualheight]{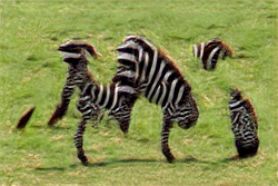}&
    \includegraphics[align=c,height=\qualheight]{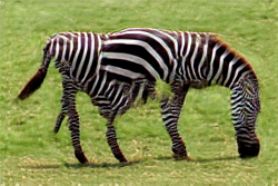}&
    \includegraphics[align=c,height=\qualheight]{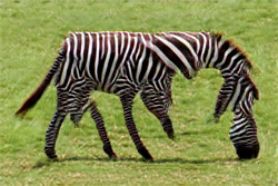}&
    \includegraphics[align=c,height=\qualheight]{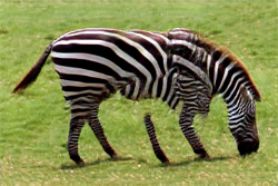}\\
    
    & \includegraphics[align=c,height=\qualheight]{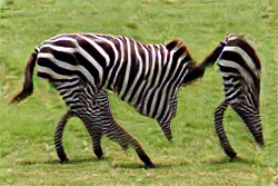}&
    \includegraphics[align=c,height=\qualheight]{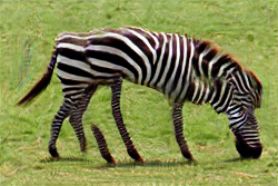}&
    \includegraphics[align=c,height=\qualheight]{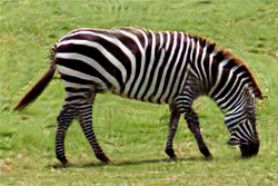}&
    \includegraphics[align=c,height=\qualheight]{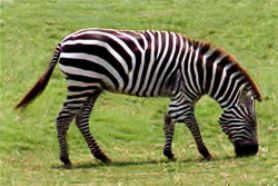}&
    \includegraphics[align=c,height=\qualheight]{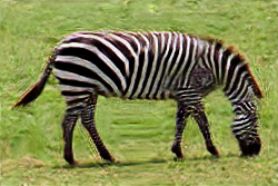}&&
    
    \includegraphics[align=c,height=\qualheight]{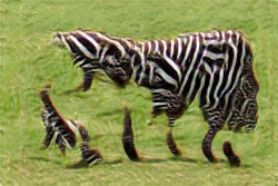}&
    \includegraphics[align=c,height=\qualheight]{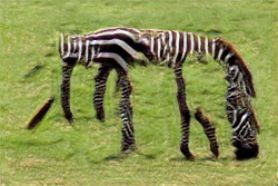}&
    \includegraphics[align=c,height=\qualheight]{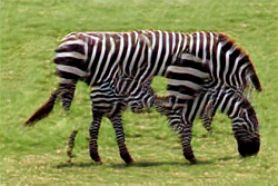}&
    \includegraphics[align=c,height=\qualheight]{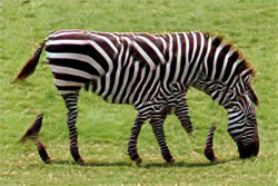}&
    \includegraphics[align=c,height=\qualheight]{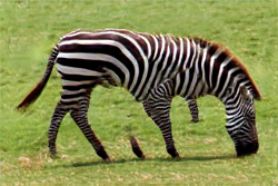}\\

    \end{tabular}
    \begin{tabular}{c*{5}{c@{\hspace{1px}}}p{2px}*{5}{c@{\hspace{1px}}}}
    \multirow{3}[10]{*}{\rotatebox[origin=c]{90}{\fsv Old Rescaling}} &
    \includegraphics[align=c,height=\qualheight]{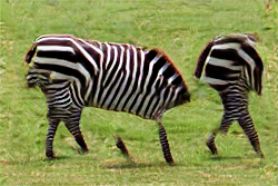}&
    \includegraphics[align=c,height=\qualheight]{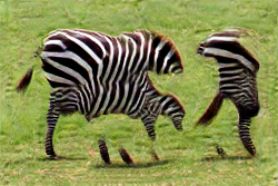}&
    \includegraphics[align=c,height=\qualheight]{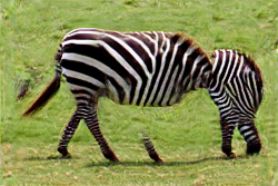}&
    \includegraphics[align=c,height=\qualheight]{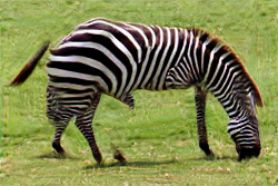}&
    \includegraphics[align=c,height=\qualheight]{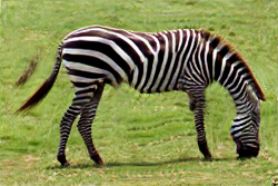}&&
    
    \includegraphics[align=c,height=\qualheight]{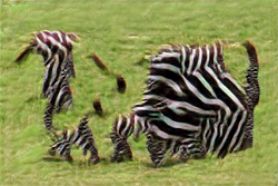}&
    \includegraphics[align=c,height=\qualheight]{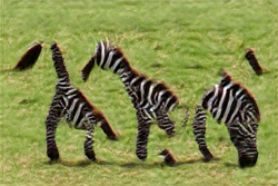}&
    \includegraphics[align=c,height=\qualheight]{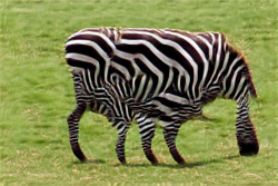}&
    \includegraphics[align=c,height=\qualheight]{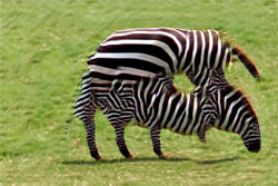}&
    \includegraphics[align=c,height=\qualheight]{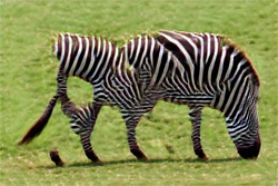}\\
    
    & \includegraphics[align=c,height=\qualheight]{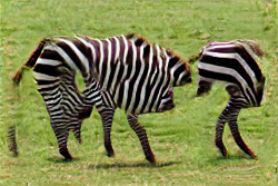}&
    \includegraphics[align=c,height=\qualheight]{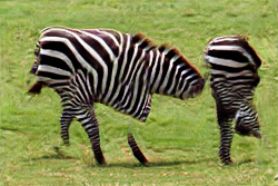}&
    \includegraphics[align=c,height=\qualheight]{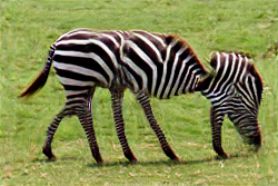}&
    \includegraphics[align=c,height=\qualheight]{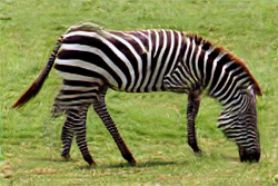}&
    \includegraphics[align=c,height=\qualheight]{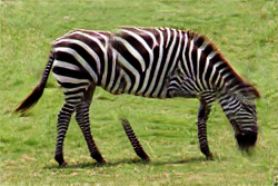}&&
    
    \includegraphics[align=c,height=\qualheight]{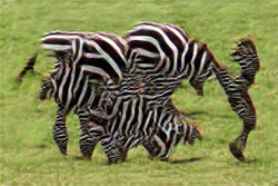}&
    \includegraphics[align=c,height=\qualheight]{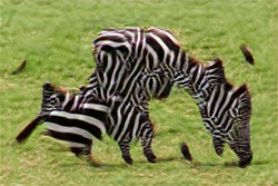}&
    \includegraphics[align=c,height=\qualheight]{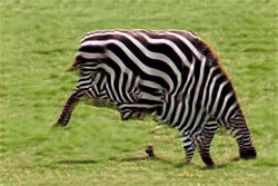}&
    \includegraphics[align=c,height=\qualheight]{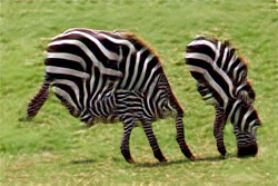}&
    \includegraphics[align=c,height=\qualheight]{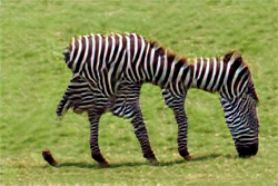}\\
    
    & \includegraphics[align=c,height=\qualheight]{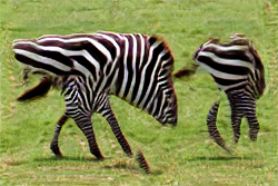}&
    \includegraphics[align=c,height=\qualheight]{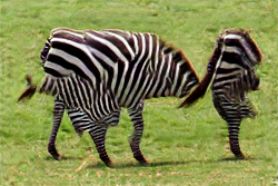}&
    \includegraphics[align=c,height=\qualheight]{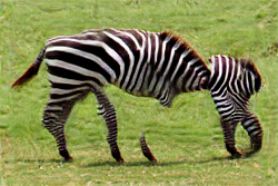}&
    \includegraphics[align=c,height=\qualheight]{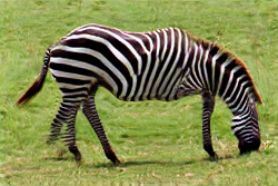}&
    \includegraphics[align=c,height=\qualheight]{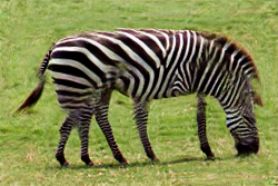}&&
    
    \includegraphics[align=c,height=\qualheight]{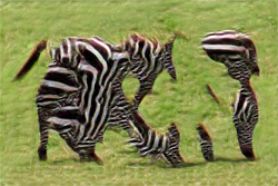}&
    \includegraphics[align=c,height=\qualheight]{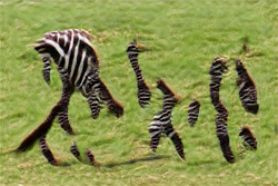}&
    \includegraphics[align=c,height=\qualheight]{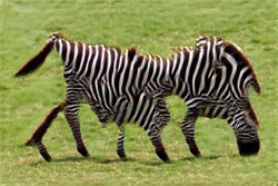}&
    \includegraphics[align=c,height=\qualheight]{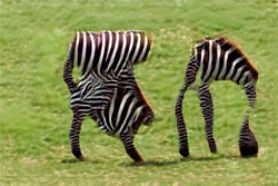}&
    \includegraphics[align=c,height=\qualheight]{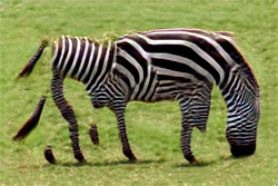}\\

    \end{tabular}
    
    \vspace{0.5em}
    \caption{Comparison of the effect of the number of trained stages and rescaling method during training. Images are randomly selected.}
    \label{fig:unconditional:rescaling:2}
\end{figure*}

\begin{table*}
    \vspace{1em}
    \centering
    \begin{tabular}{l c c c c c}
    \toprule
        Model & Confusion $\uparrow$ & SIFID $\downarrow$ & Train Time & \# Stages & \# Paramers  \\
        \midrule
        ConSinGAN & $16.0\%\pm 1.4\%$ & $0.06\pm 0.03$ & 24 min & 5.9 & {\raise.17ex\hbox{$\scriptstyle\sim$}}660,000 \\
        SinGAN & $17.0\%\pm 1.5\%$ & $0.09\pm 0.07$ & 152 min & 9.7 & {\raise.17ex\hbox{$\scriptstyle\sim$}}1,340,000 \\
        \bottomrule
    \end{tabular}
    \vspace{0.3em}
    \caption{Results of our user study and SIFID on images from the \textbf{Places} dataset.}
    \label{tab:user_study:places}
\end{table*}

\subsection{Unconditional Image Generation}
\vspace{\reduceheight}
Since our architecture is completely convolutional we can change the size of the input noise vector to generate images of various resolutions at test time.
\autoref{fig:unconditional:consingan} shows an overview of results from our method on a set of challenging images that require the generation of \emph{global} structures for the images to seem realistic. 
We observe that ConSinGAN is successfully able to capture these global structures, even if we modify the image resolution at test time.
For example, in the Stonehenge example, we can see how ``stones'' are added when the image width is increased and ``layers'' are added to the aqueduct image when the image height is increased. 

\myparagraph{Ablation}
We further examine the interplay between the learning rate scaling and the number of concurrently trained stages (\autoref{fig:overfitting}) and evaluate how varying the learning rate scaling $\delta$ (\autoref{sec:methodology}) affects training (\autoref{fig:unconditional:delta}).
As we can see in \autoref{fig:overfitting}, training with a $\delta=0.1$ leads to diverse images for most settings, with the diversity slightly decreasing with a larger number of concurrently trained stages.
When training with $\delta = 0.5$ we observe a large decrease in image diversity even when only training two stages concurrently.
As such, the number of concurrently trained stages and the learning rate scaling $\delta$ offer a trade-off between diversity and fidelity of the generated images.

\autoref{fig:unconditional:delta} visualizes how the variance in the generated images increases with decreasing $\delta$ for a model with three concurrently trained stages.
For example, when we look at the top left example (Marina Bay Sands), we observe that for a $\delta=0.5$ the overall layout of the image stays the same, with minor variations in, e.g., the appearance of the towers.
However, with a $\delta=0.1$, the appearance of the towers changes more drastically and sometimes even additional towers are added to the generated image.
Unless otherwise mentioned, all illustrated examples and all images used for the user study where generated by models for which we trained three stages concurrently with $\delta=0.1$.

\begin{figure*}[t]
    \centering
    \renewcommand{\arraystretch}{0.25}
    \def\qualheighta{1.34cm}
    \def\qualheightb{1.1926cm}
    \def\qualheightc{1.2663cm}
    \def\qualheightd{1.4003cm}
    \def\fs{\footnotesize}
    \begin{tabular}{*{2}{c@{\hspace{2px}}}p{0px}*{3}{c@{\hspace{2px}}}p{0px}*{2}{c@{\hspace{2px}}}}
    
    &&& \multicolumn{3}{c}{\fs SinGAN \cite{shaham2019singan}} && \multicolumn{2}{c}{\fs ConSinGAN} \\
    
    \fs Original & \fs Naive && \multicolumn{2}{c}{\fs 8-10 Stages} & \fs 3 Stages && \fs 3 Stages & \fs Fine-tuned \\

    \includegraphics[align=c,height=\qualheighta]{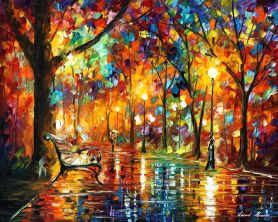}&
    \includegraphics[align=c,height=\qualheighta]{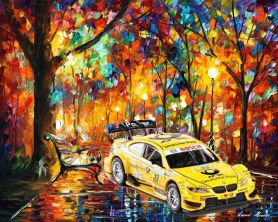}&&
    \includegraphics[align=c,height=\qualheighta]{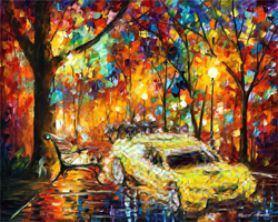}&
    \includegraphics[align=c,height=\qualheighta]{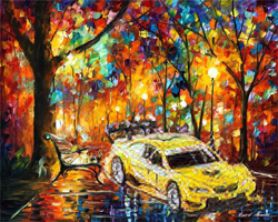}&
    \includegraphics[align=c,height=\qualheighta]{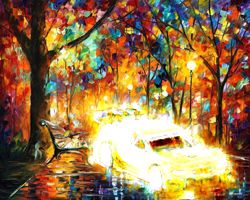}&&
    \includegraphics[align=c,height=\qualheighta]{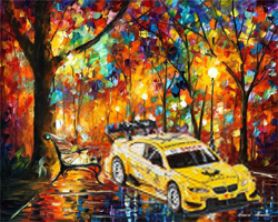}&
    \includegraphics[align=c,height=\qualheighta]{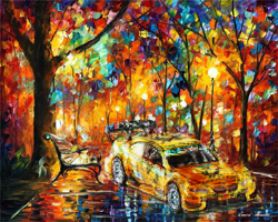} \\
    
    \includegraphics[align=c,height=\qualheightb]{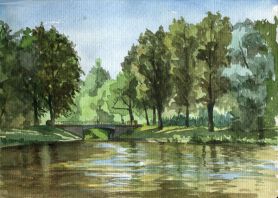}&
    \includegraphics[align=c,height=\qualheightb]{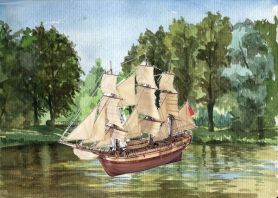}&&
    \includegraphics[align=c,height=\qualheightb]{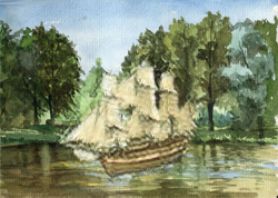}&
    \includegraphics[align=c,height=\qualheightb]{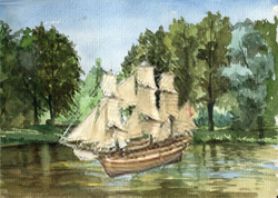}&
    \includegraphics[align=c,height=\qualheightb]{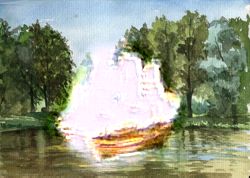}&&
    \includegraphics[align=c,height=\qualheightb]{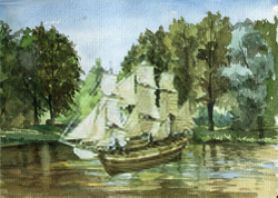}&
    \includegraphics[align=c,height=\qualheightb]{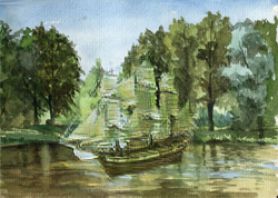} \\
    
    \includegraphics[align=c,height=\qualheightc]{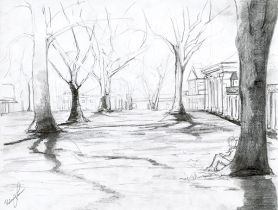}&
    \includegraphics[align=c,height=\qualheightc]{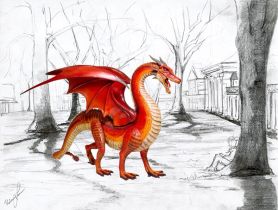}&&
    \includegraphics[align=c,height=\qualheightc]{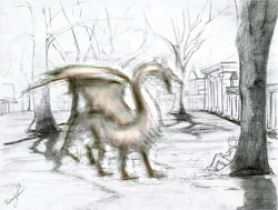}&
    \includegraphics[align=c,height=\qualheightc]{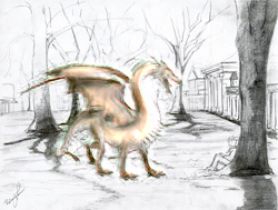}&
    \includegraphics[align=c,height=\qualheightc]{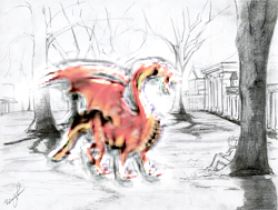}&&
    \includegraphics[align=c,height=\qualheightc]{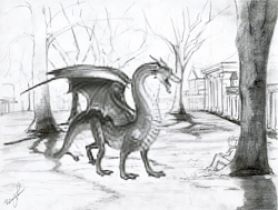}&
    \includegraphics[align=c,height=\qualheightc]{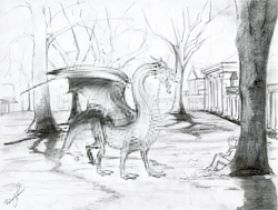} \\
    
    \includegraphics[align=c,height=\qualheightd]{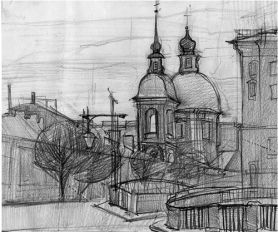}&
    \includegraphics[align=c,height=\qualheightd]{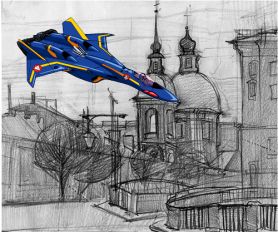}&&
    \includegraphics[align=c,height=\qualheightd]{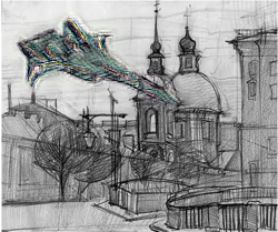}&
    \includegraphics[align=c,height=\qualheightd]{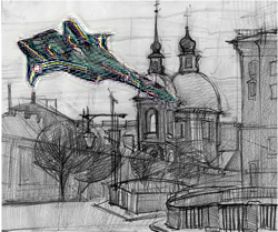}&
    \includegraphics[align=c,height=\qualheightd]{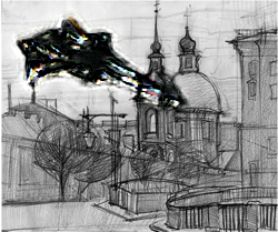}&&
    \includegraphics[align=c,height=\qualheightd]{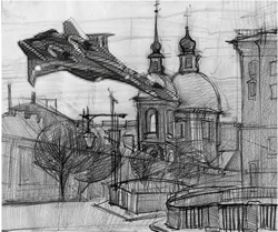}&
    \includegraphics[align=c,height=\qualheightd]{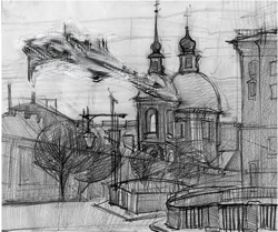} \\
    
    \end{tabular}
    \vspace{0.5em}
    \caption{Image harmonization with SinGAN and ConSinGAN}
    \label{fig:harmonization:singan}
\end{figure*}

\myparagraph{Baseline comparisons} 
We compare our model to the SinGAN~\cite{shaham2019singan} model in \autoref{fig:unconditional:consingan:singan}.
For SinGAN, we show the results of both the default rescaling method (8-10 stages) and our rescaling method (5-6 stages).
In the first example we observe that SinGAN struggles to model recurring structures (faces) in the generated images.
In the second example we observe a loss of global structure independent of the number of stages trained.
Our multi-stage training helps ensure a more consistent global structure.

\autoref{fig:unconditional:rescaling:2} further highlights the advantages of our approach by showing a detailed comparison of the images each model generates after being trained with the new or old rescaling technique.
Each column depicts three randomly sampled images from each model.
We can see the positive effect of the rescaling technique for both models, regardless of the number of trained stages.
Furthermore, we can see that our model retains better global coherence in both cases.

\begin{table*}
    \centering
    \begin{tabular}{l l l c c c c}
    \toprule
        Model & Random $\uparrow$ & Paired $\uparrow$ & SIFID $\downarrow$ & Train Time & \#  Stages & \# Parameters  \\
        \midrule
        ConSinGAN & $56.7\%\pm 1.9\%$ & $63.1\%\pm 1.8\%$ & $0.11\pm 0.06$ & 20 min & 5.9 & {\raise.17ex\hbox{$\scriptstyle\sim$}}660K \\
        SinGAN & $43.3\%\pm 1.9\%$ & $36.9\%\pm 1.8\%$ & $0.23 \pm 0.15$ & 135 min & 9.1 & {\raise.17ex\hbox{$\scriptstyle\sim$}}1.0M \\
        \bottomrule
    \end{tabular}
    \vspace{0.3em}
    \caption{Results of our user studies and SIFID on images from the \textbf{LSUN} dataset.}
    \label{tab:user_study:lsun}
\end{table*}

\myparagraph{Quantitative evaluation}
The Fr\'echet Inception Distance (FID) \cite{heusel2017gans} compares the distribution of a pre-trained network's activations between a sets of generated and real images.
The Single Image FID (SIFID) is an adaptation of the FID to the single image domain and compares the statistics of the network's activations between two individual images (generated and real).
In our experiments, we found that SIFID exhibits very high variance across different images (scores range from $1e-06$ to $1e01$) without a clear distinction of which was ``better'' or ``worse''.
In this work, we focus mostly on qualitative analyses and user studies for our evaluation but also report SIFID for comparison.

We performed quantitative evaluations on two datasets.
The first dataset is the same as the one used by SinGAN, consisting of 50 images from several categories of the `Places' dataset \cite{zhou2014learning}.
However, many of these images do not exhibit a global layout or structure.
Therefore, we also construct a second dataset, where we take five random samples from each of the ten classes of the LSUN dataset \cite{yu15lsun}.
This dataset contains classes such as ``church'' and ``bridge'' which exhibit more global structures.
We train both the SinGAN model and our model for each of the 50 images in both datasets and use the results for our evaluation.

\begin{figure*}[t]
    \centering
    \renewcommand{\arraystretch}{1}
    \def\qualheighta{1.475cm}
    \def\qualheightb{1.8087cm}
    \def\qualheightc{1.9cm}
    \def\qualheightd{2.2cm}
    \def\fs{\footnotesize}
    \begin{tabular}{*{2}{c@{\hspace{5px}}}p{0px}*{1}{c@{\hspace{5px}}}p{0px}*{2}{c@{\hspace{5px}}}}
    
    &&&&& \multicolumn{2}{c}{\fs ConSinGAN} \\[-1ex]
    
    \fs Original & \fs Naive && \fs DPH \cite{luan2018deep} && \fs 4 Stages & \fs Fine-tuned \\
    
    \includegraphics[align=c,height=\qualheighta]{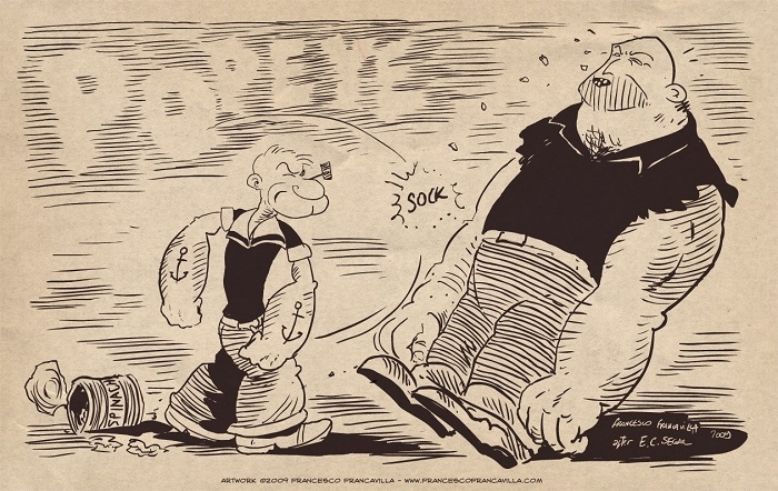}&
    \includegraphics[align=c,height=\qualheighta]{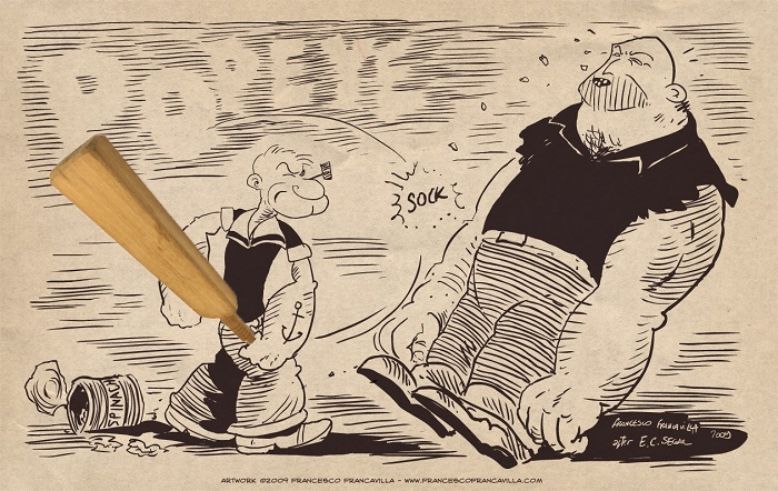}&&
    \includegraphics[align=c,height=\qualheighta]{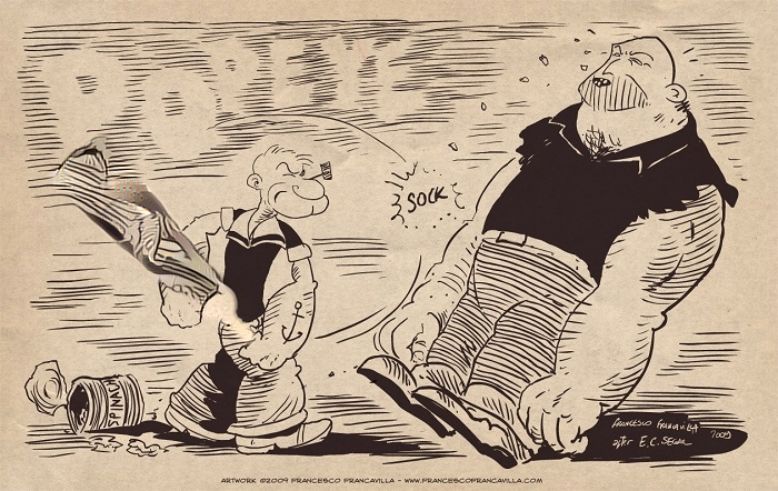}&&
    \includegraphics[align=c,height=\qualheighta]{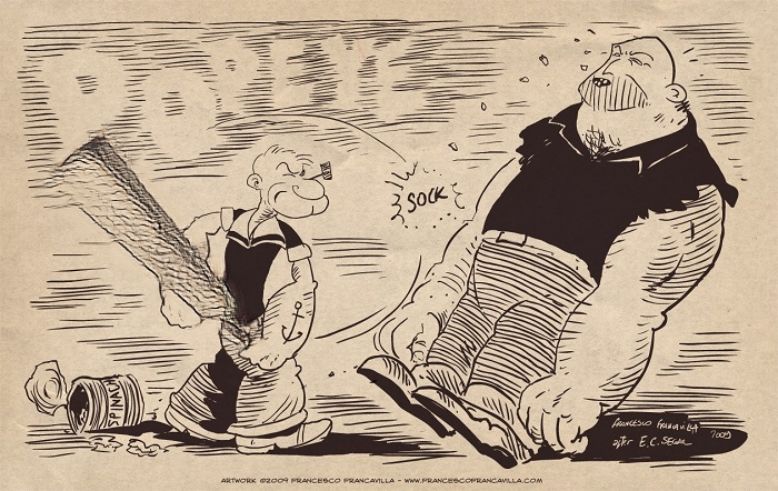}&
    \includegraphics[align=c,height=\qualheighta]{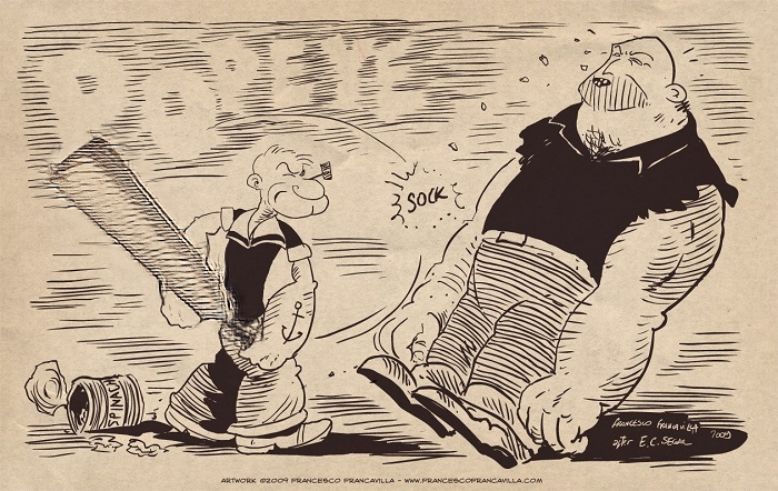} \\

    \includegraphics[clip,trim=0 70 0 90, align=c,height=\qualheightc]{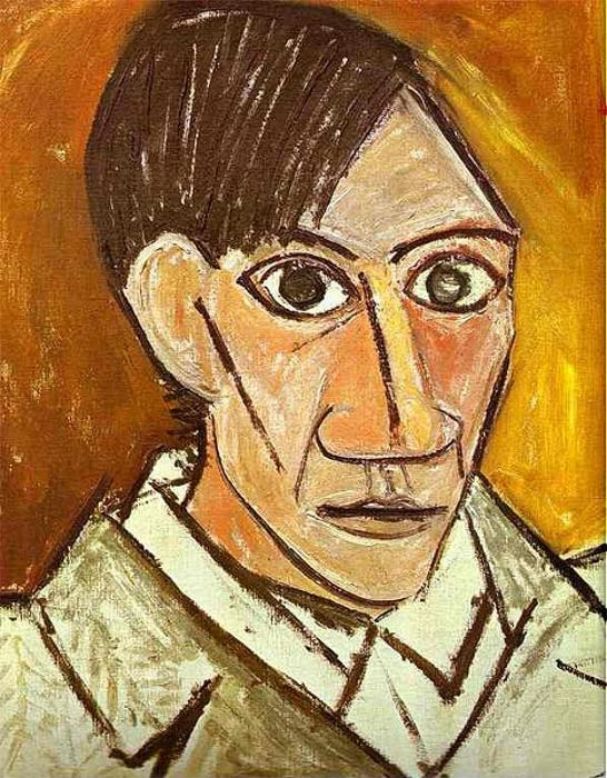}&
    \includegraphics[clip,trim=0 70 0 90, align=c,height=\qualheightc]{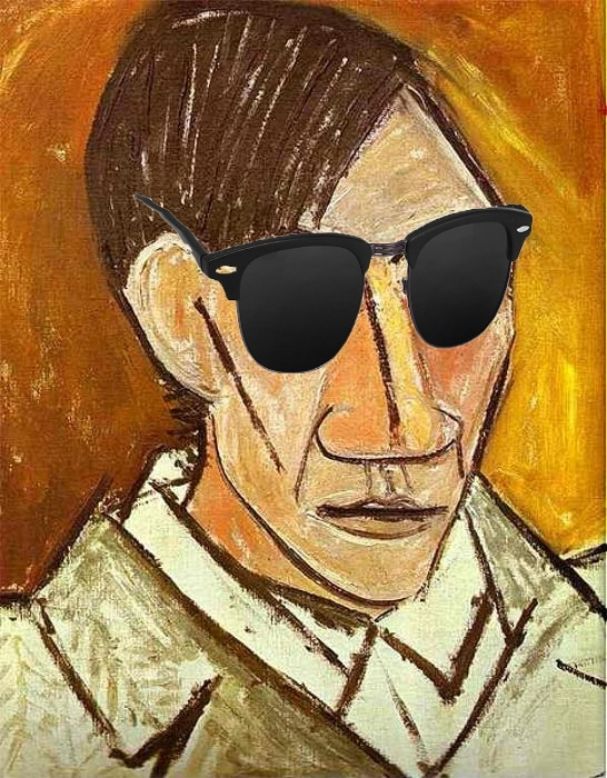}&&
    \includegraphics[clip,trim=0 70 0 90, align=c,height=\qualheightc]{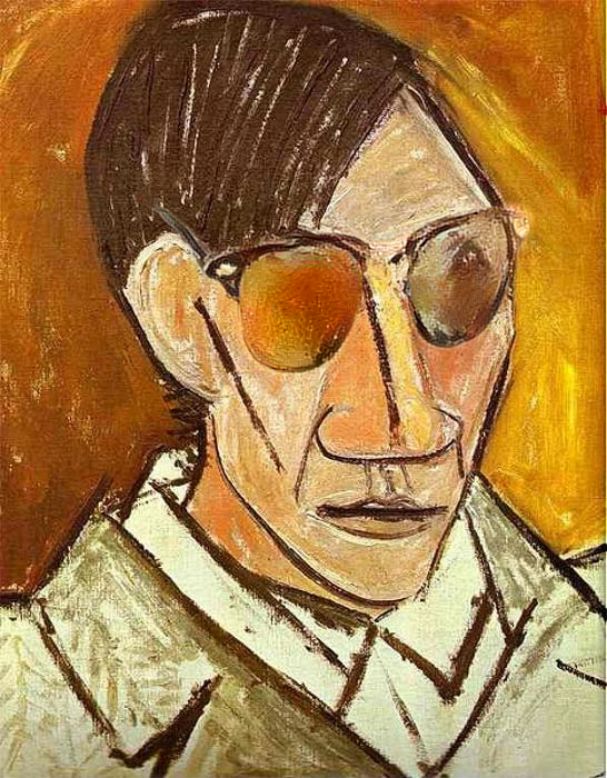}&&
    \includegraphics[clip,trim=0 70 0 90, align=c,height=\qualheightc]{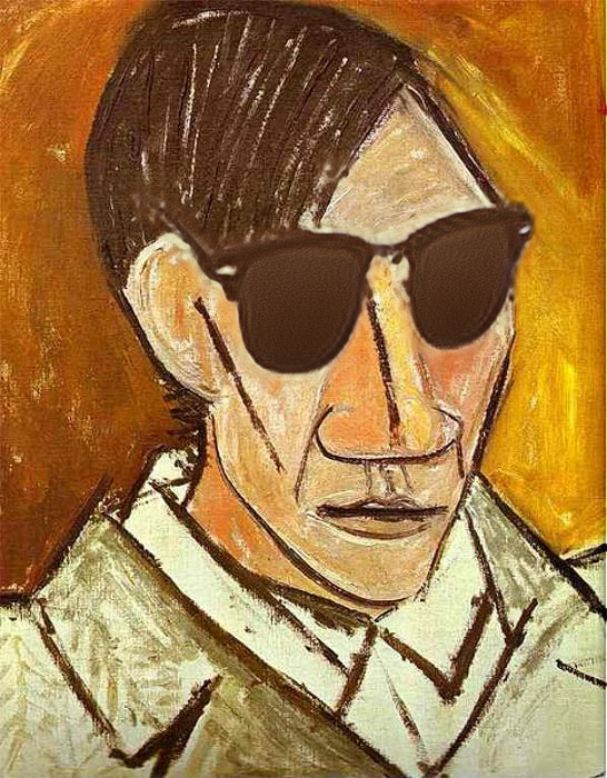}&
    \includegraphics[clip,trim=0 70 0 90, align=c,height=\qualheightc]{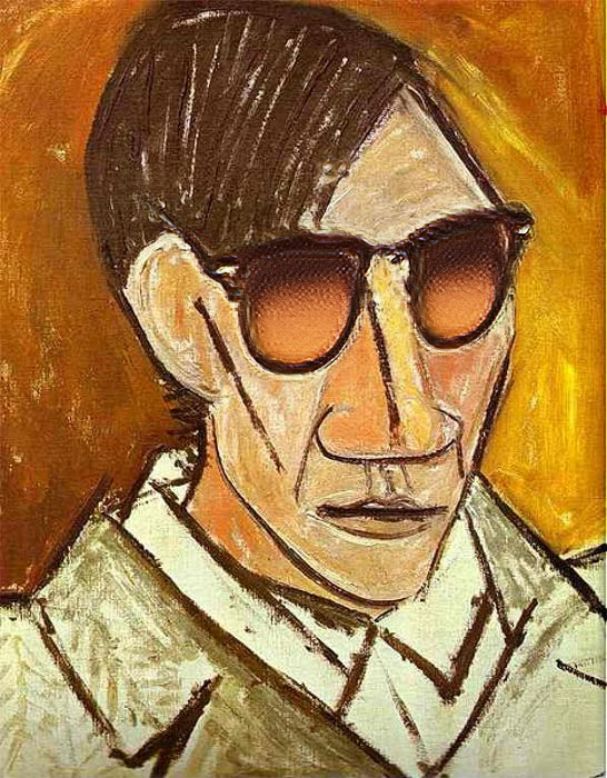} \\
    
    \includegraphics[clip,trim=0 70 0 90, align=c,height=\qualheightd]{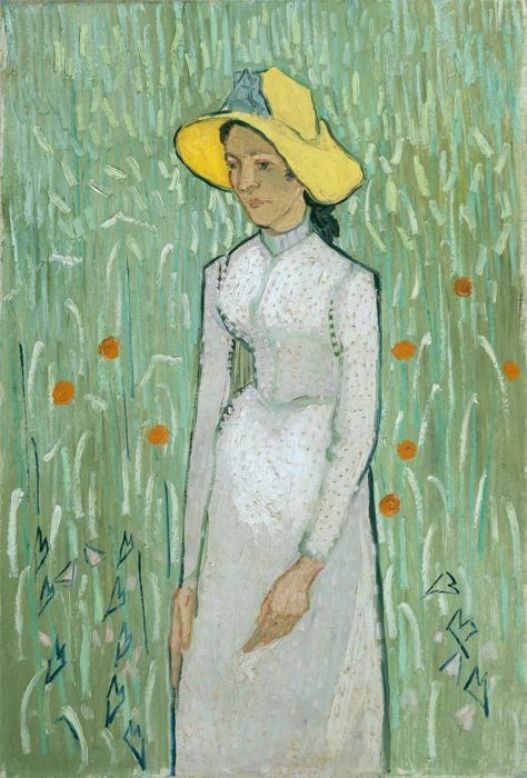}&
    \includegraphics[clip,trim=0 70 0 90, align=c,height=\qualheightd]{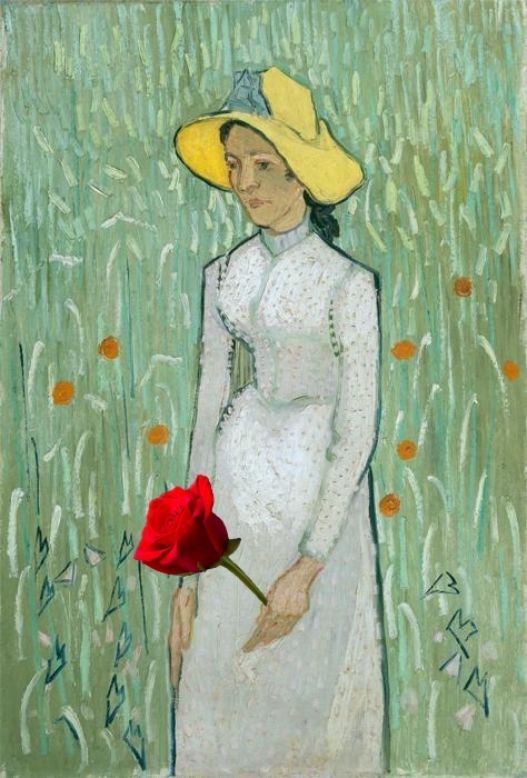}&&
    \includegraphics[clip,trim=0 70 0 90, align=c,height=\qualheightd]{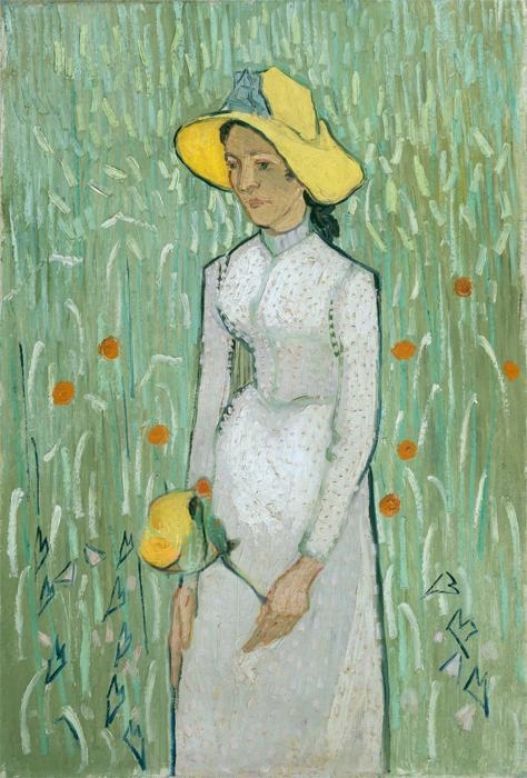}&&
    \includegraphics[clip,trim=0 70 0 90, align=c,height=\qualheightd]{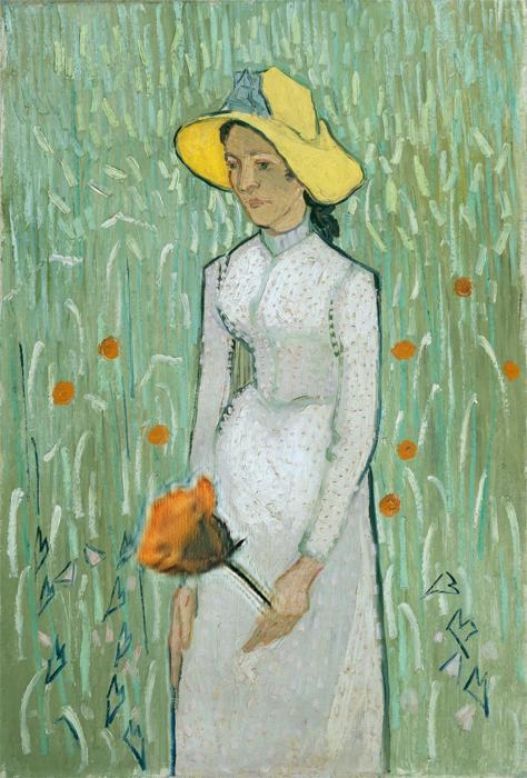}&
    \includegraphics[clip,trim=0 70 0 90, align=c,height=\qualheightd]{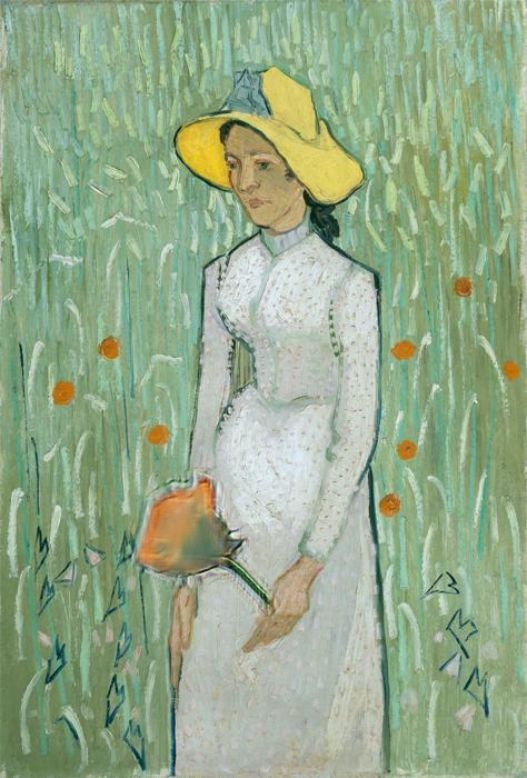} \\

    \end{tabular}
    \vspace{0.5em}
    \caption{Image harmonization comparison with Deep Painterly Harmonization (DPH) and ConSinGAN on high resolution images.}
    \label{fig:harmonization:dph:1}
\end{figure*}

\paragraph{Image Diversity}
We evaluate the diversity in our images compared to the original SinGAN model using the same measure as SinGAN: for a given training image we calculate the average of the standard deviation of all pixel values along the channel axis of 100 generated images.
Then, we normalize this value by the standard deviation of the pixel values in the training image.
On the data from the `Places' dataset, SinGAN obtains a diversity score of $0.52$, while our model's diversity is similar with a score of $0.50$.
When we increase the learning rate on lower stages by setting $\delta=0.5$ instead of the default $\delta=0.1$ we observe a lower diversity score of $0.43$ as the model learns a more precise representation of the training image (\autoref{fig:unconditional:delta}).
On the LSUN data, SinGAN obtains a much higher diversity score of $0.64$.
This is due to the fact that it often fails to model the global structure and the resulting generated images differ greatly from the training image.
Our model, on the other hand, obtains a diversity score of $0.54$ which is similar to the score on the `Places' dataset and indicates that our model can indeed learn the global structure of complex images.

\paragraph{User Study: `Places'}
We follow the same evaluation procedure as previous work \cite{isola2017image,shaham2019singan,zhang2016colorful} to compare our model with SinGAN on the same training images that were used previously in \cite{shaham2019singan}.
Users were shown our generated image and its respective training image for one second each and were asked to identify the real image.
We reproduced the user study from the SinGAN paper with our own trained SinGAN and ConSinGAN models.
As we can see in \autoref{tab:user_study:places} our model achieves results similar to the SinGAN model.
However, our model is trained on fewer stages and with fewer parameters and obtains a better SIFID score of $0.06$, compared to SinGAN's $0.09$.
Furthermore, the images generated by ConSinGAN often still exhibit a better global structure, but one second is not enough time for users to identify this.

\paragraph{User Study: `LSUN'}
Since the images from the LSUN dataset are much more challenging than the images from the `Places' dataset we do not compare the generated images against the real images, but instead compare the images generated by SinGAN to the ones generated by ConSinGAN.
We generate 10 images per training image, resulting in 500 generated images each from SinGAN and ConSinGAN, and use these to compare the models in two different user studies.

In both versions, the participants see the two images generated by the two models next to each other and need to judge which image is better.
We do not enforce a time limit, so participants can look at both images for as long as they choose.
The difference between the two versions of the user study is how we sample the generated images.
In the first version (``random'') we randomly sample one image from the set of generated images of SinGAN and ConSinGAN each.
This means that the two images likely come from different classes (e.g.\ `church' vs. `conference room').
In the second version (``paired'') we sample two images that were generated from the same training image.
We perform both user studies using Amazon Mechanical Turk, with 50 participants comparing 60 pairs of images for each study.

\autoref{tab:user_study:lsun} shows how often users picked images generated by a given model for each of the two settings.
We see that users prefer the images generated by ConSinGAN in both settings and that, again, our model achieves a better SIFID.
This is the case even though our model only trains on six stages, has fewer parameters than SinGAN, and takes less time to train.
The images from LSUN vary in difficulty and global structure.
This might explain why our model performs even better in the paired setting since this setting guarantees that we always compare the two models on images of the same difficulty.
Overall, our experiments show that ConSinGAN allows for the generation of more believable images, especially when they exhibit some degree of global structure, with less training time and a smaller model than SinGAN.

\subsection{Image Harmonization}
\vspace{\reduceheight}
We now show results on image harmonization examples and compare our model to SinGAN and Deep Painterly Harmonization \cite{luan2018deep} for high-resolution images.

\myparagraph{Training Details}
We train ConSinGAN with the same hyperparameters for all images without any fine-tuning of hyperparameters for the different images.
The general architecture is the same as for unconditional image generation, however, we only train the model for exactly three stages per image.
We train for 1,000 iterations per stage and randomly sample from different data augmentation techniques to obtain a ``new'' training image at each iteration as described in \autoref{sec:methodology}.
When we fine-tune a model on a given specific image we use a model trained on the general style image and use the target image directly as input (instead of the style image with random augmentation transformations) to train the model for an additional 500 iterations.

\myparagraph{Comparison with SinGAN}
\autoref{fig:harmonization:singan} shows comparisons between SinGAN and ConSinGAN.
The first two columns show the original images we trained on and the naive cut-and-paste images that are the input to our trained model at test time.
The next three images show the results of a trained SinGAN model, where the first two are the results of a fully trained model.
We insert the naive image at all stages of the model and choose the two best results, while the third image is the result when we train SinGAN on only three stages.
The final two columns show the results of the ConSinGAN.
Training ConSinGAN takes less than 10 minutes for a given image when the coarse side of the image has a resolution of 250 pixels.
Fine-tuning a model on a specific image takes roughly 2-3 minutes.
Training SinGAN takes roughly 120 minutes as before, since we need to train the full model, even if only some of the later stages are used at test time.

We see that ConSinGAN performs similar to or better than SinGAN, even though we only train ConSinGAN for 3 stages.
ConSinGAN also generally introduces fewer artifacts into the harmonized image, while SinGAN often changes the surface structure of the added objects.
See for example the first row in \autoref{fig:harmonization:singan}, where SinGAN adds artifacts onto the car, while ConSinGAN keeps the original objects consistent.
When we fine-tune the ConSinGAN model on specific images we can get even more interesting results, as, e.g., the car gets absorbed much more into the colors of the overall background.
The bottom two rows of \autoref{fig:harmonization:singan} show results when we add colorful objects to black-and-white paintings.
When training SinGAN on only three stages like ConSinGAN it usually fails completely to harmonize the objects at test time.
Even the images harmonized after training SinGAN on 8-10 stages often contain some of the original colors, while ConSinGAN manages to completely transfer the objects to black-and-white versions.
Again, further fine-tuning ConSinGAN on the specific images leads to an even stronger ``absorption'' of the objects.

\myparagraph{Comparison with DPH}
\autoref{fig:harmonization:dph:1} shows comparisons between ConSinGAN, adapted to harmonize high-resolution images, and Deep Painterly Harmonization (DPH) \cite{luan2018deep}.
The images have a resolution of roughly 700 pixels on the longer side, as opposed to the 250 pixels used by the SinGAN examples.
In order to produce these high-resolution images, we add another stage to our ConSinGAN architecture, i.e.\ we now train four stages, and training time increases to roughly 30-40 minutes per image.
This is in contrast to many style-transfer approaches and also DPH, which have additional hyperparameters such as the style and content weight which need to be fine-tuned for a specific style image.

We can see that the outputs of ConSinGAN usually differ from the outputs of DPH, but are still realistic and visually pleasing.
This is even the case when our model has never seen the naive copy-and-paste image at train time, but only uses it at test time.
In contrast to this, DPH requires as input the style input, the naive copy-and-past input, and the mask which specifies the location of the copied object in the image.
Again, fine-tuning our model sometimes leads to even better results, but even the model trained only with random image augmentations performs well.
While our training time is quite long, we only need to train our model once for a given image and can then add different objects at different locations at test time.
This is not possible with DPH, which needs to be retrained whenever the copied object changes.

\section{Conclusion}
\vspace{\reduceheight}

We introduced ConSinGAN, a GAN inspired by a number of best practices discovered for training single-image GANs. 
Our model is trained on sequentially increasing image resolutions, to first learn the global structure of the image, before learning texture and stylistic details later.
Compared to other models, our
approach allows for control over how closely the internal patch distribution of the training image is learned by adjusting the number of concurrently trained stages and the learning rate scaling at lower stages. 
Through this, we can decide how much diversity we want in the generated images.
We also introduce a new image rescaling approach that allows training on fewer image scales than before.
We show that our approach can be trained on a single image and can be used for tasks such as unconditional image generation, harmonization, editing, and animation while being smaller and more efficient to train than previous models.

\myparagraph{Acknowledgements}
\noindent The authors gratefully acknowledge partial support from the German Research Foundation DFG under project CML (TRR 169).

\clearpage
{\small
\bibliographystyle{ieee_fullname}
\bibliography{egbib}

\begin{thebibliography}{10}\itemsep=-1pt

\bibitem{asano2020critical}
Yuki~M Asano, Christian Rupprecht, and Andrea Vedaldi.
\newblock A critical analysis of self-supervision, or what we can learn from a
  single image.
\newblock In {\em International Conference on Learning Representations}, 2020.

\bibitem{bagon2008good}
Shai Bagon, Oren Boiman, and Michal Irani.
\newblock What is a good image segment? a unified approach to segment
  extraction.
\newblock In {\em European Conference on Computer Vision}, pages 30--44.
  Springer, 2008.

\bibitem{bell2019blind}
Sefi Bell-Kligler, Assaf Shocher, and Michal Irani.
\newblock Blind super-resolution kernel estimation using an internal-gan.
\newblock In {\em Advances in Neural Information Processing Systems}, pages
  284--293, 2019.

\bibitem{benaim2020structural}
Sagie Benaim, Ron Mokady, Amit Bermano, Daniel Cohen-Or, and Lior Wolf.
\newblock Structural-analogy from a single image pair.
\newblock {\em arXiv preprint arXiv:2004.02222}, 2020.

\bibitem{bergmann2017learning}
Urs Bergmann, Nikolay Jetchev, and Roland Vollgraf.
\newblock Learning texture manifolds with the periodic spatial gan.
\newblock In {\em International Conference on Machine Learning}, pages
  469--477, 2017.

\bibitem{brock2018large}
Andrew Brock, Jeff Donahue, and Karen Simonyan.
\newblock Large scale gan training for high fidelity natural image synthesis.
\newblock In {\em International Conference on Learning Representations}, 2019.

\bibitem{cho2008patch}
Taeg~Sang Cho, Moshe Butman, Shai Avidan, and William~T Freeman.
\newblock The patch transform and its applications to image editing.
\newblock In {\em Proceedings of the IEEE Computer Vision and Pattern
  Recognition}, pages 1--8. IEEE, 2008.

\bibitem{dekel2015revealing}
Tali Dekel, Tomer Michaeli, Michal Irani, and William~T Freeman.
\newblock Revealing and modifying non-local variations in a single image.
\newblock {\em ACM Transactions on Graphics (TOG)}, 34(6):1--11, 2015.

\bibitem{demir2018patch}
Ugur Demir and Gozde Unal.
\newblock Patch-based image inpainting with generative adversarial networks.
\newblock {\em arXiv preprint arXiv:1803.07422}, 2018.

\bibitem{gandelsman2019double}
Yossi Gandelsman, Assaf Shocher, and Michal Irani.
\newblock Double-dip": Unsupervised image decomposition via coupled
  deep-image-priors.
\newblock In {\em Proceedings of the IEEE Computer Vision and Pattern
  Recognition}, volume~6, page~2, 2019.

\bibitem{glasner2009super}
Daniel Glasner, Shai Bagon, and Michal Irani.
\newblock Super-resolution from a single image.
\newblock In {\em Proceedings of the IEEE International Conference on Computer
  Vision}, pages 349--356. IEEE, 2009.

\bibitem{goodfellow2014generative}
Ian Goodfellow, Jean Pouget-Abadie, Mehdi Mirza, Bing Xu, David Warde-Farley,
  Sherjil Ozair, Aaron Courville, and Yoshua Bengio.
\newblock Generative adversarial nets.
\newblock In {\em Advances in Neural Information Processing Systems}, pages
  2672--2680, 2014.

\bibitem{gulrajani2017improved}
Ishaan Gulrajani, Faruk Ahmed, Martin Arjovsky, Vincent Dumoulin, and Aaron~C
  Courville.
\newblock Improved training of wasserstein gans.
\newblock In {\em Advances in Neural Information Processing Systems}, pages
  5767--5777, 2017.

\bibitem{he2012statistics}
Kaiming He and Jian Sun.
\newblock Statistics of patch offsets for image completion.
\newblock In {\em European Conference on Computer Vision}, pages 16--29.
  Springer, 2012.

\bibitem{he2016deep}
Kaiming He, Xiangyu Zhang, Shaoqing Ren, and Jian Sun.
\newblock Deep residual learning for image recognition.
\newblock In {\em Proceedings of the IEEE Computer Vision and Pattern
  Recognition}, pages 770--778, 2016.

\bibitem{heusel2017gans}
Martin Heusel, Hubert Ramsauer, Thomas Unterthiner, Bernhard Nessler, and Sepp
  Hochreiter.
\newblock Gans trained by a two time-scale update rule converge to a local nash
  equilibrium.
\newblock In {\em Advances in Neural Information Processing Systems}, pages
  6626--6637, 2017.

\bibitem{hinz2019generating}
Tobias Hinz, Stefan Heinrich, and Stefan Wermter.
\newblock Generating multiple objects at spatially distinct locations.
\newblock In {\em International Conference on Learning Representations}, 2019.

\bibitem{huang2015single}
Jia-Bin Huang, Abhishek Singh, and Narendra Ahuja.
\newblock Single image super-resolution from transformed self-exemplars.
\newblock In {\em Proceedings of the IEEE Computer Vision and Pattern
  Recognition}, pages 5197--5206, 2015.

\bibitem{isola2017image}
Phillip Isola, Jun-Yan Zhu, Tinghui Zhou, and Alexei~A Efros.
\newblock Image-to-image translation with conditional adversarial networks.
\newblock In {\em Proceedings of the IEEE Computer Vision and Pattern
  Recognition}, pages 1125--1134, 2017.

\bibitem{jetchev2016texture}
Nikolay Jetchev, Urs Bergmann, and Roland Vollgraf.
\newblock Texture synthesis with spatial generative adversarial networks.
\newblock {\em arXiv preprint arXiv:1611.08207}, 2016.

\bibitem{Karnewar_2019_CVPR}
Animesh Karnewar and Oliver Wang.
\newblock Msg-gan: Multi-scale gradients for generative adversarial networks.
\newblock In {\em Proceedings of the IEEE Computer Vision and Pattern
  Recognition}, 2020.

\bibitem{karras2017progressive}
Tero Karras, Timo Aila, Samuli Laine, and Jaakko Lehtinen.
\newblock Progressive growing of gans for improved quality, stability, and
  variation.
\newblock In {\em International Conference on Learning Representations}, 2018.

\bibitem{karras2019style}
Tero Karras, Samuli Laine, and Timo Aila.
\newblock A style-based generator architecture for generative adversarial
  networks.
\newblock In {\em Proceedings of the IEEE Computer Vision and Pattern
  Recognition}, pages 4401--4410, 2019.

\bibitem{karras2019analyzing}
Tero Karras, Samuli Laine, Miika Aittala, Janne Hellsten, Jaakko Lehtinen, and
  Timo Aila.
\newblock Analyzing and improving the image quality of stylegan.
\newblock In {\em Proceedings of the IEEE Computer Vision and Pattern
  Recognition}, 2020.

\bibitem{li2016precomputed}
Chuan Li and Michael Wand.
\newblock Precomputed real-time texture synthesis with markovian generative
  adversarial networks.
\newblock In {\em European Conference on Computer Vision}, pages 702--716.
  Springer, 2016.

\bibitem{lin2020tuigan}
Jianxin Lin, Yingxue Pang, Yingce Xia, Zhibo Chen, and Jiebo Luo.
\newblock Tuigan: Learning versatile image-to-image translation with two
  unpaired images.
\newblock {\em European Conference on Computer Vision}, 2020.

\bibitem{luan2018deep}
Fujun Luan, Sylvain Paris, Eli Shechtman, and Kavita Bala.
\newblock Deep painterly harmonization.
\newblock {\em Computer Graphics Forum}, 37(4):95--106, 2018.

\bibitem{mao2019program}
Jiayuan Mao, Xiuming Zhang, Yikai Li, William~T Freeman, Joshua~B Tenenbaum,
  and Jiajun Wu.
\newblock Program-guided image manipulators.
\newblock In {\em Proceedings of the IEEE International Conference on Computer
  Vision}, pages 4030--4039, 2019.

\bibitem{mastan2019multi}
Indra~Deep Mastan and Shanmuganathan Raman.
\newblock Multi-level encoder-decoder architectures for image restoration.
\newblock In {\em IEEE/CVF Conference on Computer Vision and Pattern
  Recognition Workshops}, 2019.

\bibitem{mastan2020dcil}
Indra~Deep Mastan and Shanmuganathan Raman.
\newblock Dcil: Deep contextual internal learning for image restoration and
  image retargeting.
\newblock In {\em The IEEE Winter Conference on Applications of Computer
  Vision}, pages 2366--2375, 2020.

\bibitem{mechrez2019saliency}
Roey Mechrez, Eli Shechtman, and Lihi Zelnik-Manor.
\newblock Saliency driven image manipulation.
\newblock {\em Machine Vision and Applications}, 30(2):189--202, 2019.

\bibitem{michaeli2014blind}
Tomer Michaeli and Michal Irani.
\newblock Blind deblurring using internal patch recurrence.
\newblock In {\em European Conference on Computer Vision}, pages 783--798.
  Springer, 2014.

\bibitem{shaham2019singan}
Tamar~Rott Shaham, Tali Dekel, and Tomer Michaeli.
\newblock Singan: Learning a generative model from a single natural image.
\newblock In {\em Proceedings of the IEEE International Conference on Computer
  Vision}, pages 4570--4580, 2019.

\bibitem{Shocher_2019_ICCV}
Assaf Shocher, Shai Bagon, Phillip Isola, and Michal Irani.
\newblock Ingan: Capturing and retargeting the "dna" of a natural image.
\newblock In {\em Proceedings of the IEEE International Conference on Computer
  Vision}, 2019.

\bibitem{shocher2018zero}
Assaf Shocher, Nadav Cohen, and Michal Irani.
\newblock “zero-shot” super-resolution using deep internal learning.
\newblock In {\em Proceedings of the IEEE Computer Vision and Pattern
  Recognition}, pages 3118--3126, 2018.

\bibitem{simakov2008summarizing}
Denis Simakov, Yaron Caspi, Eli Shechtman, and Michal Irani.
\newblock Summarizing visual data using bidirectional similarity.
\newblock In {\em Proceedings of the IEEE Computer Vision and Pattern
  Recognition}, pages 1--8. IEEE, 2008.

\bibitem{tlusty2018modifying}
Tal Tlusty, Tomer Michaeli, Tali Dekel, and Lihi Zelnik-Manor.
\newblock Modifying non-local variations across multiple views.
\newblock In {\em Proceedings of the IEEE Computer Vision and Pattern
  Recognition}, pages 6276--6285, 2018.

\bibitem{ulyanov2018deep}
Dmitry Ulyanov, Andrea Vedaldi, and Victor Lempitsky.
\newblock Deep image prior.
\newblock In {\em Proceedings of the IEEE Computer Vision and Pattern
  Recognition}, pages 9446--9454, 2018.

\bibitem{vinker2020deep}
Yael Vinker, Eliahu Horwitz, Nir Zabari, and Yedid Hoshen.
\newblock Deep single image manipulation.
\newblock {\em arXiv preprint arXiv:2007.01289}, 2020.

\bibitem{yang2019deep}
Wenming Yang, Xuechen Zhang, Yapeng Tian, Wei Wang, Jing-Hao Xue, and Qingmin
  Liao.
\newblock Deep learning for single image super-resolution: A brief review.
\newblock {\em IEEE Transactions on Multimedia}, 21(12):3106--3121, 2019.

\bibitem{yu15lsun}
Fisher Yu, Yinda Zhang, Shuran Song, Ari Seff, and Jianxiong Xiao.
\newblock Lsun: Construction of a large-scale image dataset using deep learning
  with humans in the loop.
\newblock {\em arXiv preprint arXiv:1506.03365}, 2015.

\bibitem{zhang2019internal}
Haotian Zhang, Long Mai, Ning Xu, Zhaowen Wang, John Collomosse, and Hailin
  Jin.
\newblock An internal learning approach to video inpainting.
\newblock In {\em Proceedings of the IEEE International Conference on Computer
  Vision}, pages 2720--2729, 2019.

\bibitem{zhang2016colorful}
Richard Zhang, Phillip Isola, and Alexei~A Efros.
\newblock Colorful image colorization.
\newblock In {\em European Conference on Computer Vision}, pages 649--666.
  Springer, 2016.

\bibitem{zhou2014learning}
Bolei Zhou, Agata Lapedriza, Jianxiong Xiao, Antonio Torralba, and Aude Oliva.
\newblock Learning deep features for scene recognition using places database.
\newblock In {\em Advances in Neural Information Processing Systems}, pages
  487--495, 2014.

\bibitem{zhou2018non}
Yang Zhou, Zhen Zhu, Xiang Bai, Dani Lischinski, Daniel Cohen-Or, and Hui
  Huang.
\newblock Non-stationary texture synthesis by adversarial expansion.
\newblock {\em ACM Transactions on Graphics (TOG)}, 37(4):1--13, 2018.

\bibitem{zhu2017unpaired}
Jun-Yan Zhu, Taesung Park, Phillip Isola, and Alexei~A Efros.
\newblock Unpaired image-to-image translation using cycle-consistent
  adversarial networks.
\newblock In {\em Proceedings of the IEEE International Conference on Computer
  Vision}, pages 2223--2232, 2017.

\bibitem{zhu2017toward}
Jun-Yan Zhu, Richard Zhang, Deepak Pathak, Trevor Darrell, Alexei~A Efros,
  Oliver Wang, and Eli Shechtman.
\newblock Toward multimodal image-to-image translation.
\newblock In {\em Advances in Neural Information Processing Systems}, pages
  465--476, 2017.

\bibitem{zontak2011internal}
Maria Zontak and Michal Irani.
\newblock Internal statistics of a single natural image.
\newblock In {\em Proceedings of the IEEE Computer Vision and Pattern
  Recognition}, pages 977--984. IEEE, 2011.

\bibitem{zontak2013separating}
Maria Zontak, Inbar Mosseri, and Michal Irani.
\newblock Separating signal from noise using patch recurrence across scales.
\newblock In {\em Proceedings of the IEEE Computer Vision and Pattern
  Recognition}, pages 1195--1202, 2013.

\end{thebibliography}
}
\end{document}